\documentclass[a4paper,conference]{IEEEtran}
\usepackage[latin9]{inputenc}
\usepackage{booktabs}
\usepackage{url}
\usepackage{multirow}
\usepackage{amsmath}
\usepackage{amssymb}
\usepackage{graphicx}
\usepackage{microtype}
\usepackage[ruled,vlined]{algorithm2e}

\makeatletter

\pdfpageheight\paperheight
\pdfpagewidth\paperwidth

\providecommand{\tabularnewline}{\\}

\usepackage{comment}
\usepackage{color}
\usepackage{array}
\usepackage{multirow}
\usepackage[noadjust]{cite}

\ifCLASSINFOpdf
\else
\fi
\makeatother

\begin{document}
\title{Towards Robust Learning with Different Label Noise Distributions}
\author{\IEEEauthorblockN{Diego Ortego, Eric Arazo, Paul Albert, Noel E. O'Connor, Kevin
McGuinness}\IEEEauthorblockA{Insight Centre for Data Analytics, Dublin City University (DCU)}\IEEEauthorblockA{\{diego.ortego, eric.arazo\}@insight-centre.org}}
\maketitle
\begin{abstract}
Noisy labels are an unavoidable consequence of labeling processes
and detecting them is an important step towards preventing performance
degradations in Convolutional Neural Networks. Discarding noisy labels
avoids a harmful memorization, while the associated image content
can still be exploited in a semi-supervised learning (SSL) setup.
Clean samples are usually identified using the small loss trick, i.e.
they exhibit a low loss. However, we show that different noise distributions
make the application of this trick less straightforward and propose
to continuously relabel all images to reveal a discriminative loss
against multiple distributions. SSL is then applied twice, once to
improve the clean-noisy detection and again for training the final
model. We design an experimental setup based on ImageNet32/64 for
better understanding the consequences of representation learning with
differing label noise distributions and find that non-uniform out-of-distribution
noise better resembles real-world noise and that in most cases intermediate
features are not affected by label noise corruption. Experiments in
CIFAR-10/100, ImageNet32/64 and WebVision (real-world noise) demonstrate
that the proposed label noise Distribution Robust Pseudo-Labeling
(DRPL) approach gives substantial improvements over recent state-of-the-art.
Code is available at \url{https://git.io/JJ0PV}.
\end{abstract}

\IEEEpeerreviewmaketitle{}

\section{Introduction\label{sec:Introduction}}

Modern representation learning, i.e. the extraction of useful information
to build classifiers or other predictors \cite{2013_TPAMI_RepresentationLearning},
in computer vision is led by Convolutional Neural Networks (CNNs)
\cite{2018_CVPR_ImClassif,2016_arXiv_Homography,2016_CVPR_ActionClassif,2017_ICCV_EventCaptioning,2018_arXiv_descriptors,2016_CVPR_YOLO,2017_CVPR_PSPNet}.
Their widespread use is attributable to their capability to model
complex patterns when vast amounts of labeled data are available \cite{2019_CVPR_RevisitingSelf}.
This supervision requirement limits exploiting the vast amounts of
web images as it is infeasible to label them for each particular task.
However, leveraging these data might drive visual representation learning
a step forward.

\emph{What can we do to relax supervision}? One could adopt transfer
learning or domain adaptation \cite{2018_Neurocomputing_DomainAdaptationSurvey},
where representations from a source domain are transferred to a target
domain where fewer labels are available. This approach, however, makes
the target domain dependent on the source one. Learning representations
from scratch in the target domain, on the other hand, may lead to
better representations. Several alternatives exist: semi-supervised
learning (SSL), which jointly learns from few labeled images and extensive
unlabeled ones \cite{2019_arXiv_Pseudo,2019_NeurIPS_MixMatch}; self-supervised
learning, where data provides the supervision \cite{2019_CVPR_RevisitingSelf,2017_CVPR_SelfSeg};
or learning with label noise, where automatic labeling processes introduce
noise in the observed labels \cite{2018_CVPR_JointOpt,2019_CVPR_JointOptimizImproved}.
This paper focuses on this last alternative, which has attracted much
recent interest \cite{2019_ICML_BynamicBootstrapping,2019_NeurIPS_LDMI,2019_AAAI_Safeguard,2019_CVPR_JointOptimizImproved}.

Learning with label noise is challenging; recent studies on the generalization
capabilities of deep networks~\cite{2017_ICLR_Rethinking} demonstrate
that noisy labels are easily fit by CNNs, harming generalization.
There is, however, a key observation on how CNNs memorize corruptions:
they tend to learn easy patterns first and these patterns are closer
to clean data patterns, i.e. correctly labeled images \cite{2017_ICML_Memorization,2017_ICLR_Rethinking},
thus exhibiting lower loss than images with noisy or incorrect labels.
This phenomenon is commonly named small loss \cite{2018_CVPR_JointOpt,2019_AAAI_Safeguard,2019_CVPR_JointOptimizImproved}
and recent works exploit this \emph{small loss trick} to identify
clean and noisy samples \cite{2019_ICML_BynamicBootstrapping,2018_NeurIPS_CoTeaching,2019_ICML_SELFIE}.

Approaches dealing with label noise can be categorized into: loss
correction \cite{2019_ICML_BynamicBootstrapping,2018_ICML_MentorNet,2017_CVPR_ForwardLoss,2015_ICLR_Bootstrapping},
when the loss is weighted to correct the label noise effect; relabeling
\cite{2018_CVPR_JointOpt,2019_CVPR_JointOptimizImproved}, when observed
labels are replaced by an estimation of the true labels; and approaches
that discard noisy labels to transform the problem into SSL \cite{2018_WACV_SemiSupNoise,2019_ICCV_NegativeLearning}.
Despite the variety of approaches and comparative evaluations, it
is not clear which of them behave better. Most approaches are exhaustively
compared on synthetic label noise in CIFAR data \cite{2009_CIFAR}
and then tested in real-world datasets such as Clothing1M \cite{2015_CVPR_GraphModelNoise}
or WebVision \cite{2017_arXiv_WebVision}. However, while uniform
and non-uniform label flips are considered in CIFAR (in-distribution
noise), real-world noise also contains out-of-distribution samples
\cite{2017_arXiv_WebVision}, thus leading to a distribution mismatch
between synthetic and real data that may impact representation learning.
Despite this mismatch, when experimenting with synthetic noise the
true labels are known, thus enabling memorization analysis.

In light of these limitations, we undertake an exhaustive study on
different label noise distributions and propose a general solution
to tackle all of them. In particular, we adopt the $32\times32$ and
$64\times64$ versions of the ImageNet dataset and artificially introduce
4 different label noise types (2 in-distribution and 2 out-of-distribution).
Our analysis reveals differences in the way labels are memorized for
different distributions, thus making the application of the small
loss trick over training losses harder. We therefore propose a noise detection strategy that reveals a suitable loss to apply the small loss trick regardless of the noise distribution. We find that discarding noisy
labels with our method and training in a semi-supervised manner outperforms
all other approaches considered. Our contributions include:
\begin{itemize}
\item A framework to study label noise with different distributions providing
a more complete understanding. 
\item A label noise detection method agnostic to the noise distribution
that substantially outperforms other recent methods \cite{2018_WACV_SemiSupNoise,2019_ICCV_NegativeLearning,2019_ICML_SELFIE}.
\item A study of label noise memorization effect in representation learning,
showing that the discrimination power of intermediate representations
is not affected.
\item An extensive evaluation using multiple label noise distributions and
real-world noise, demonstrating both superior performance of our label
noise Distribution Robust Pseudo-Labeling (DRPL) approach and contributing
to a better understanding of existing methods. 
\end{itemize}

\section{Related work}

Label noise is a well-known problem in machine learning \cite{2013_TNNLS_LabelNoise}.
Recent efforts in image classification focus on dealing with in-distribution
noise \cite{2019_ICCV_NegativeLearning}, where the set of possible
labels $S$ is known and noisy labels belong to this set. However,
label noise might also come from outside the distribution \cite{2018_CVPR_IterativeNoise},
which occurs in many real-world scenarios \cite{2017_arXiv_WebVision}.

Several label noise distributions can affect dataset annotations,
namely \emph{uniform} or \emph{non-uniform} random label noise. Uniform
label noise means the true labels are flipped to a different class
with uniform random probability. Non-uniform noise has different flipping
probabilities for each class.

Loss correction approaches~\cite{2019_ICML_BynamicBootstrapping,2017_CVPR_ForwardLoss,2015_ICLR_Bootstrapping,2019_AAAI_Safeguard}
either modify the loss directly or the network probabilities to compensate
for the incorrect guidance provided by the noisy samples. \cite{2015_ICLR_Bootstrapping}
extend the loss with a perceptual term that introduces a reliance
on the model prediction. Han et al.~\cite{2019_ICCV_PrototypesBootstrapping}
extend it by adopting, as a perceptual term,  class estimations based
on sample prototypes, while Arazo et al.~\cite{2019_ICML_BynamicBootstrapping}
 dynamically weight the perceptual term based on the clean-noisy per-sample
probability given by a label noise model. Patrini et al.~\cite{2017_CVPR_ForwardLoss}
estimate the label noise transition matrix $T$, which specifies the
probability of one label being flipped to another, and correct the
softmax probability by multiplying by $T$. In the same spirit, Yao
et al. \cite{2019_AAAI_Safeguard} propose to estimate $T$ in a Bayesian
non-parametric form and deduce a dynamic label regression method to
train the classifier and model the noise. 

Other loss correction approaches reduce the contribution of noisy
samples by defining per-sample weights using a mentor network \cite{2018_ICML_MentorNet},
an unsupervised estimation of data complexity \cite{2018_ECCV_CurrNet},
or metric learning to pull noisy samples representations away from
clean ones \cite{2018_CVPR_IterativeNoise}. Robust loss functions
are studied in \cite{2018_NeurIPS_GCE}, which define the generalized
cross-entropy loss by jointly exploiting the benefits of mean absolute
error and cross-entropy losses. \cite{2019_NeurIPS_LDMI} propose
the Determinant-based Mutual Information, a generalized version of
mutual information that is provably insensitive to noise patterns
and amounts.

Other approaches relabel the noisy samples by modeling their noise
through conditional random fields \cite{2017_NIPS}, or CNNs \cite{2017_CVPR_CRFrelabel}
using a set of clean samples, which limits their applicability. Tanaka
et al.~\cite{2018_CVPR_JointOpt} and Yi et al. \cite{2019_CVPR_JointOptimizImproved}
have, however, demonstrated that it is possible to do sample relabeling
using the network predictions or estimated label distributions as
soft labels.

It has, moreover, recently been demonstrated~\cite{2018_WACV_SemiSupNoise,2019_ICCV_NegativeLearning,2020_ICLR_DivideMix}
that it is useful to discard samples that are likely to be noisy while
still using them in a semi-supervised setup. \cite{2018_WACV_SemiSupNoise}
define clean samples by inspecting prediction-label agreement, whereas
\cite{2019_ICCV_NegativeLearning} use high softmax probabilities
to distinguish clean samples after performing negative learning, i.e.
minimizing the probability of predicting likely incorrect classes.
More recently, \cite{2020_ICLR_DivideMix} jointly trains two networks
and exploits their interactions to dynamically re-define the clean
samples during training.

\section{From label noise to semi-supervised learning}

\begin{figure*}[t]
\centering{}\setlength{\tabcolsep}{0.0pt} 
\global\long\def\arraystretch{0}%
\begin{tabular}{cccc}
\includegraphics[width=0.22\textwidth]{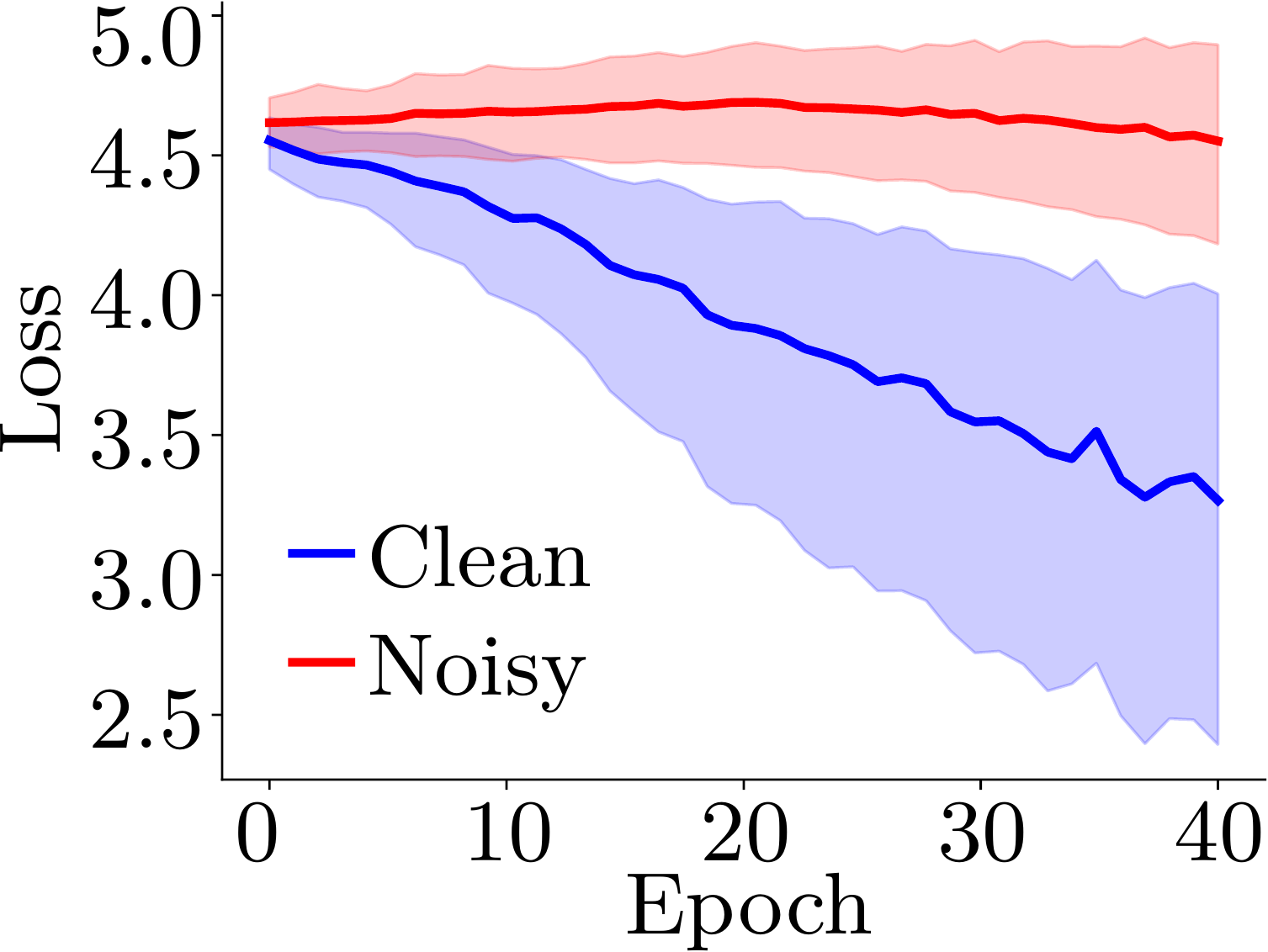} & \includegraphics[width=0.22\textwidth]{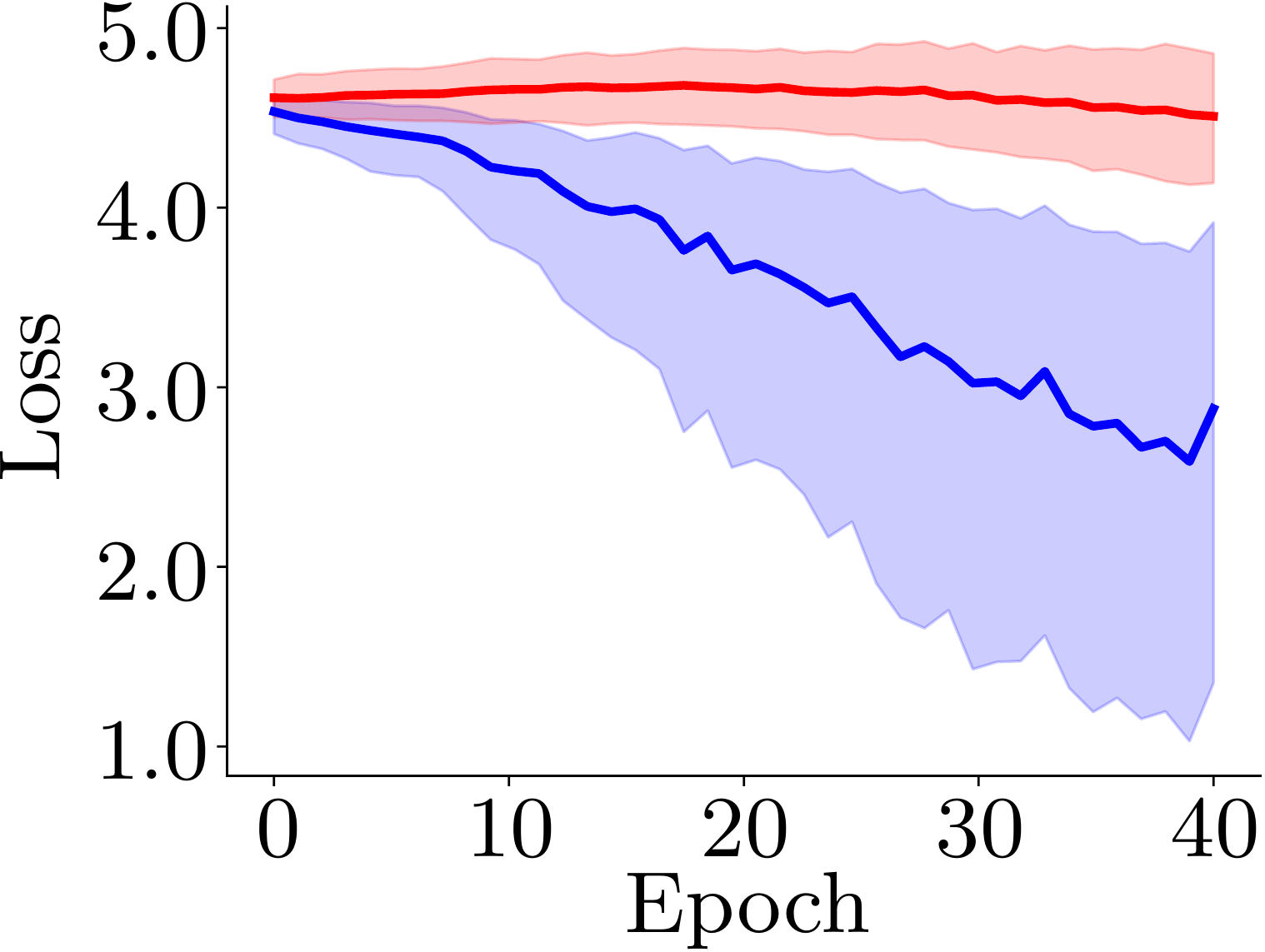} & \includegraphics[width=0.22\textwidth]{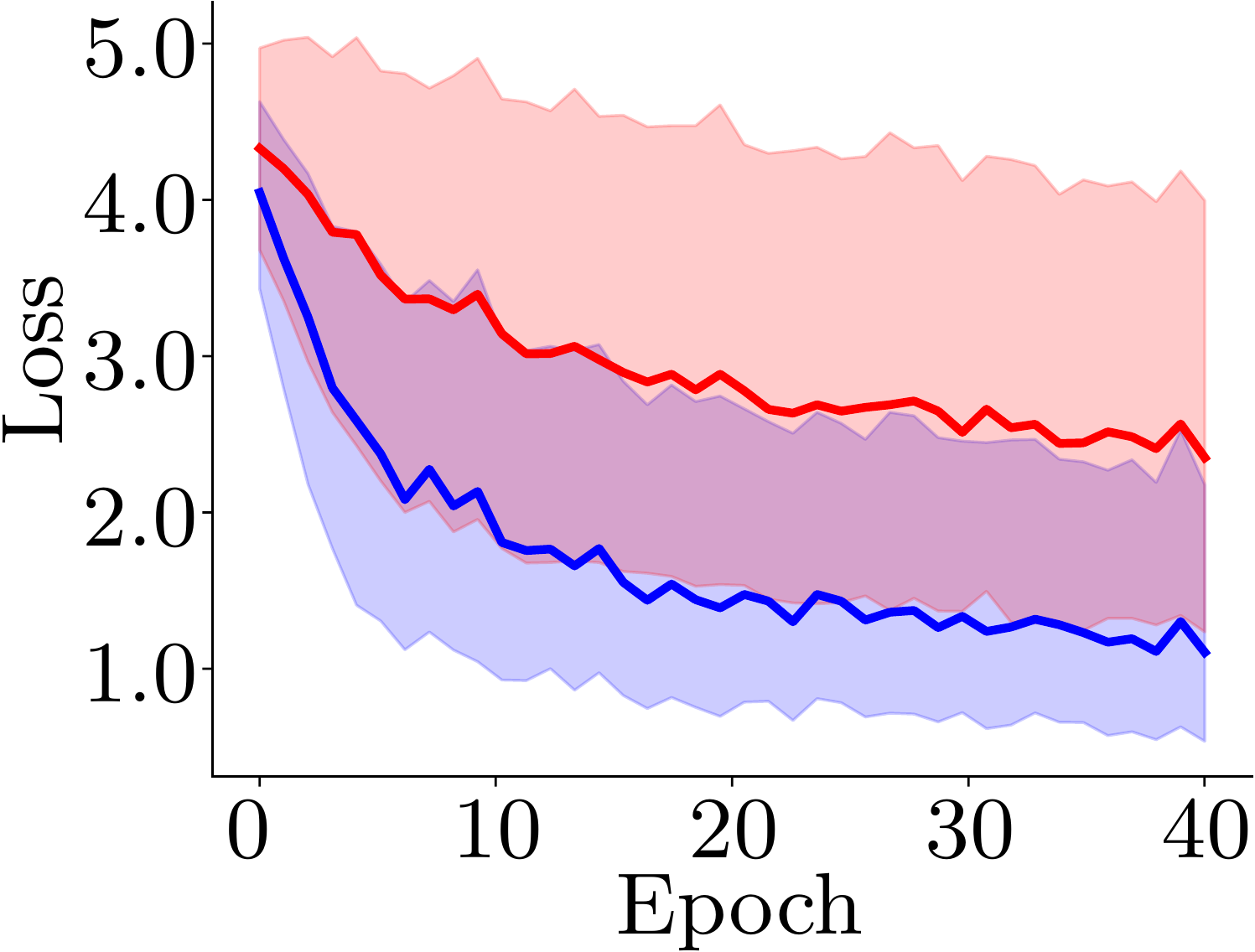} & \includegraphics[width=0.22\textwidth]{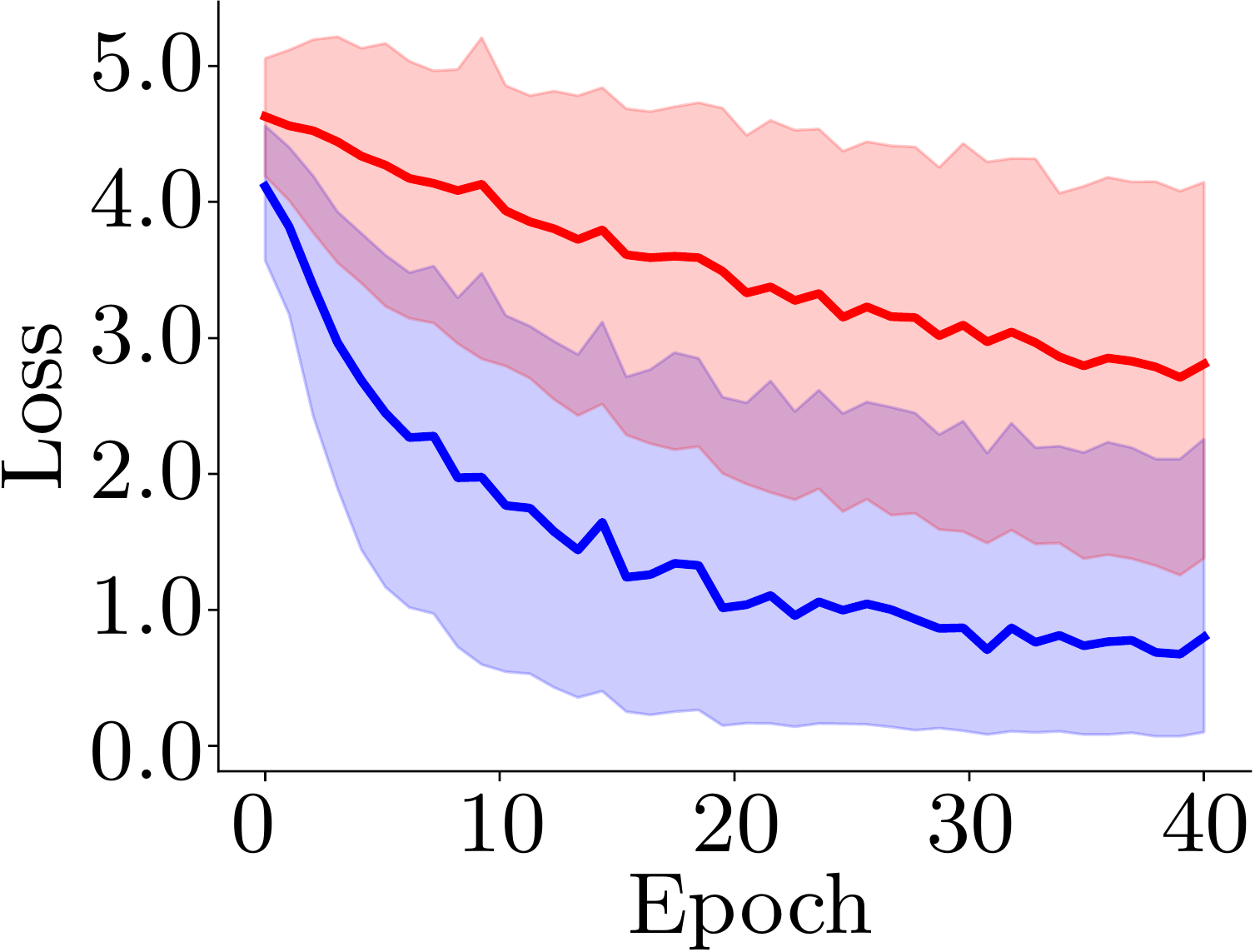}\tabularnewline
\end{tabular}\caption{\label{fig:LossNoises}Loss for clean (blue) and noisy (red) samples
for different label noise distributions in 100 classes of ImageNet32.
80\% uniform in-distribution noise (left); 80\% uniform out-of-distribution
noise (mid-left); 50\% non-uniform in-distribution noise (mid-right);
and 50\% non-uniform out-of-distribution noise (right). Training:
40 epochs with a PreAct ResNet-18 \cite{2016_ECCV_PreActResNet} with
learning rate of 0.1 and cross-entropy loss.}
\end{figure*}
Image classification can be formulated as learning a model $h_{\theta}(x)$
from a set of training examples $\mathcal{D}=\left\{ \left(x_{i},y_{i}\right)\right\} _{i=1}^{N}$
with $y_{i}\in\left\{ 0,1\right\} ^{C}$ being the one-hot encoding
true label corresponding to $x_{i}$. In our case, $h_{\theta}$ is
a CNN and $\theta$ represents the model parameters (weights and biases).
As we are considering classification under label noise, the label
$y_{i}$ can be noisy (i.e. $x_{i}$ is a noisy sample). This training
under label noise can be redefined through SSL, where the $N$ samples
are split into a set of $N_{u}$ unlabeled samples $\mathcal{D}_{u}=\left\{ x_{i}\right\} _{i=1}^{N_{u}}$
and a set of $N_{l}$ labeled samples $\mathcal{D}_{l}=\left\{ \left(x_{i},y_{i}\right)\right\} _{i=1}^{N_{l}}$.
Ideally, $\mathcal{D}_{l}$ would contain the clean samples and $\mathcal{D}_{u}$
the noisy ones. In practice, clean and noisy samples sets are unknown
and must be estimated. Detected clean (noisy) samples are used as
the labeled (unlabeled) set and, as such, detection accuracy will
influence the SSL success. The remainder of this section introduces
the proposed two-stage label noise detection (Subsections \ref{subsec:Stage1}
and \ref{subsec:Stage2}) and briefly explains the SSL approach \cite{2019_arXiv_Pseudo}
(Subsection \ref{subsec:SSLapproach}).

\subsection{Label noise detection: first stage\label{subsec:Stage1}}

Recent literature assumes that using the small loss trick when training
with cross-entropy leads to accurate clean-noisy discrimination \cite{2019_ICML_BynamicBootstrapping,2018_NeurIPS_CoTeaching,2019_ICML_SELFIE}.
However, as Figure~\ref{fig:LossNoises} shows, different label noise
distributions result in different behavior. Noisy samples tend to
have higher loss than clean samples for both in-distribution and out-of-distribution
noise but the different noise types exhibit different complexities.
The uniform noise types (Figure~\ref{fig:LossNoises} (left part))
exhibit higher separation between the losses of clean (blue) and noisy
(red) samples than the non-uniform ones (Figure~\ref{fig:LossNoises}
(right part)). This shows that a straightforward application of the
small loss trick will likely encounter some difficulties.

Based on the evidence that clean data is easier to fit than mislabeled
data \cite{2017_ICML_Memorization}, we propose to identify the clean
data by fully relabeling all samples using the network predictions
and analyzing which samples still fit the original labels. This stage
is key for the success of our approach across different label noise
distributions. For relabeling we adopt \cite{2018_CVPR_JointOpt},
which optimizes the following loss function:{\small{}{} 
\begin{equation}
\ell_{t}(\phi)=\begin{cases}
-\frac{1}{N}\sum_{i=1}^{N}y_{i}^{T}\log\left(h_{\phi}(x_{i})\right)+R & t\leq q\\
\\
-\frac{1}{N}\sum_{i=1}^{N}\tilde{y}_{i}^{T}\log\left(h_{\phi}(x_{i})\right)+R & t>q
\end{cases},\label{eq:RelabelingLoss}
\end{equation}
\begin{equation}
R=\lambda_{1}R_{H}+\lambda_{2}R_{A},\label{eq:RegJointOpt}
\end{equation}
}where $t$ indexes the epochs, $q$ defines the number of warm-up
epochs with the original (potentially noisy) labels $y$, and $R_{H}$
and $R_{A}$ are two regularization terms (see details in \cite{2018_CVPR_JointOpt})
weighted by $\lambda_{1}$ and $\lambda_{2}$ included to ensure convergence.
The first phase ($t\leq q$) trains without relabeling and uses a
high learning rate to prevent fitting the label noise, while the second
($t>q$) relabels all samples by computing soft pseudo-labels $\tilde{y}$
that are re-estimated every epoch using the softmax predictions $h_{\phi}(x)$.
For simplicity we omit index $t$ inside Eq. \eqref{eq:RelabelingLoss}.

\begin{figure*}[t]
\centering{}\setlength{\tabcolsep}{0.0pt}%
\begin{tabular}{cccc}
\includegraphics[width=0.23\textwidth]{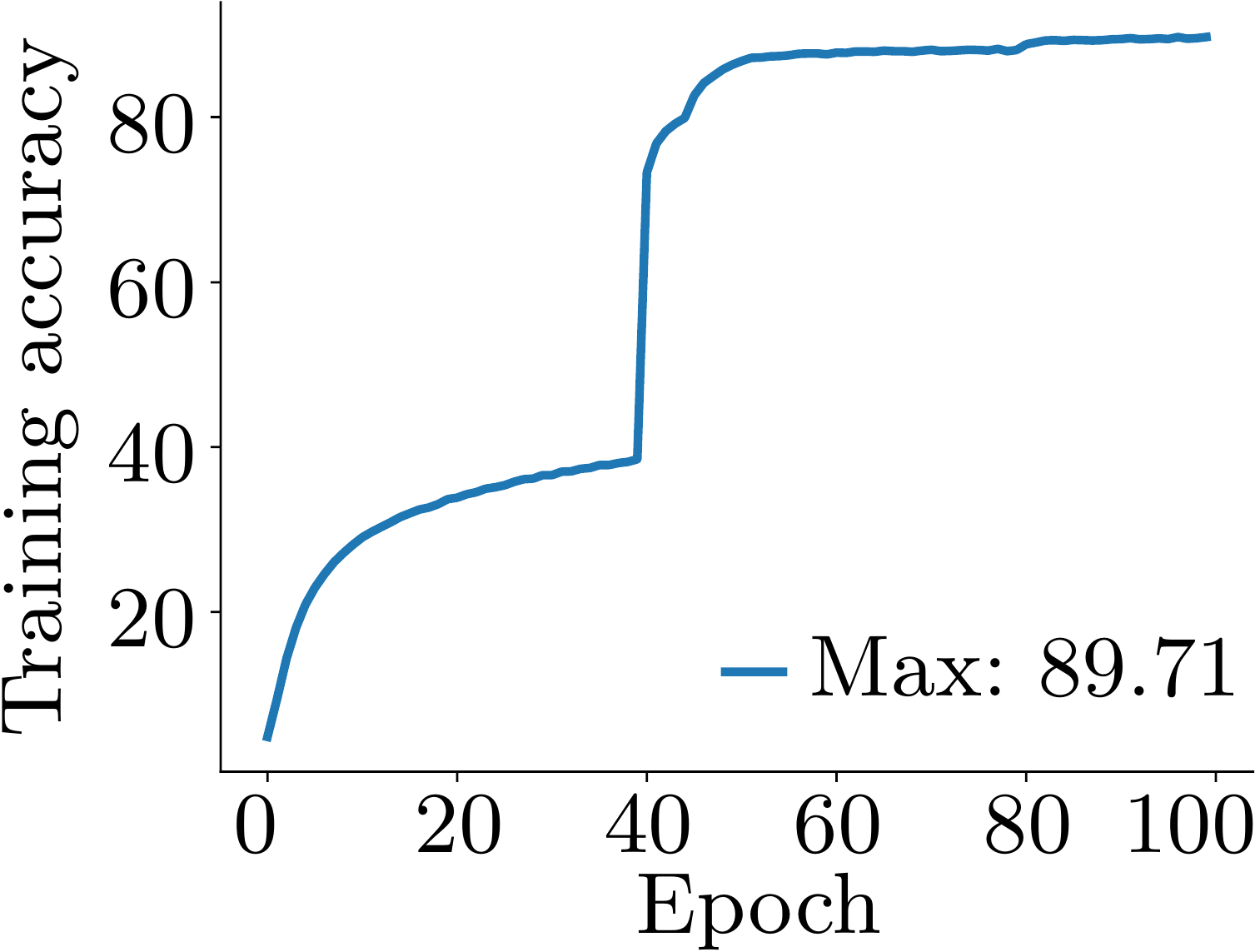}  & \includegraphics[width=0.23\textwidth]{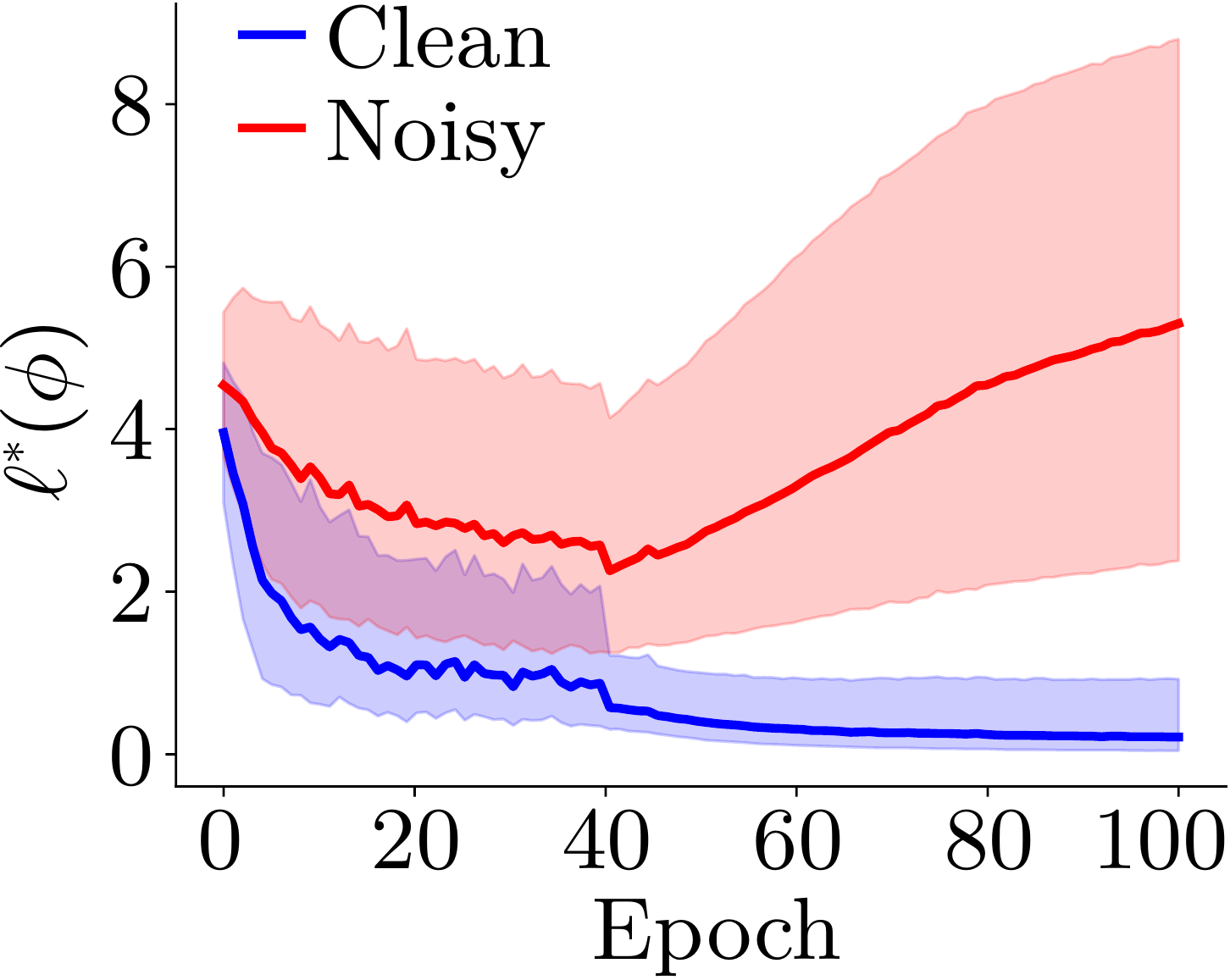}  & \includegraphics[width=0.23\textwidth]{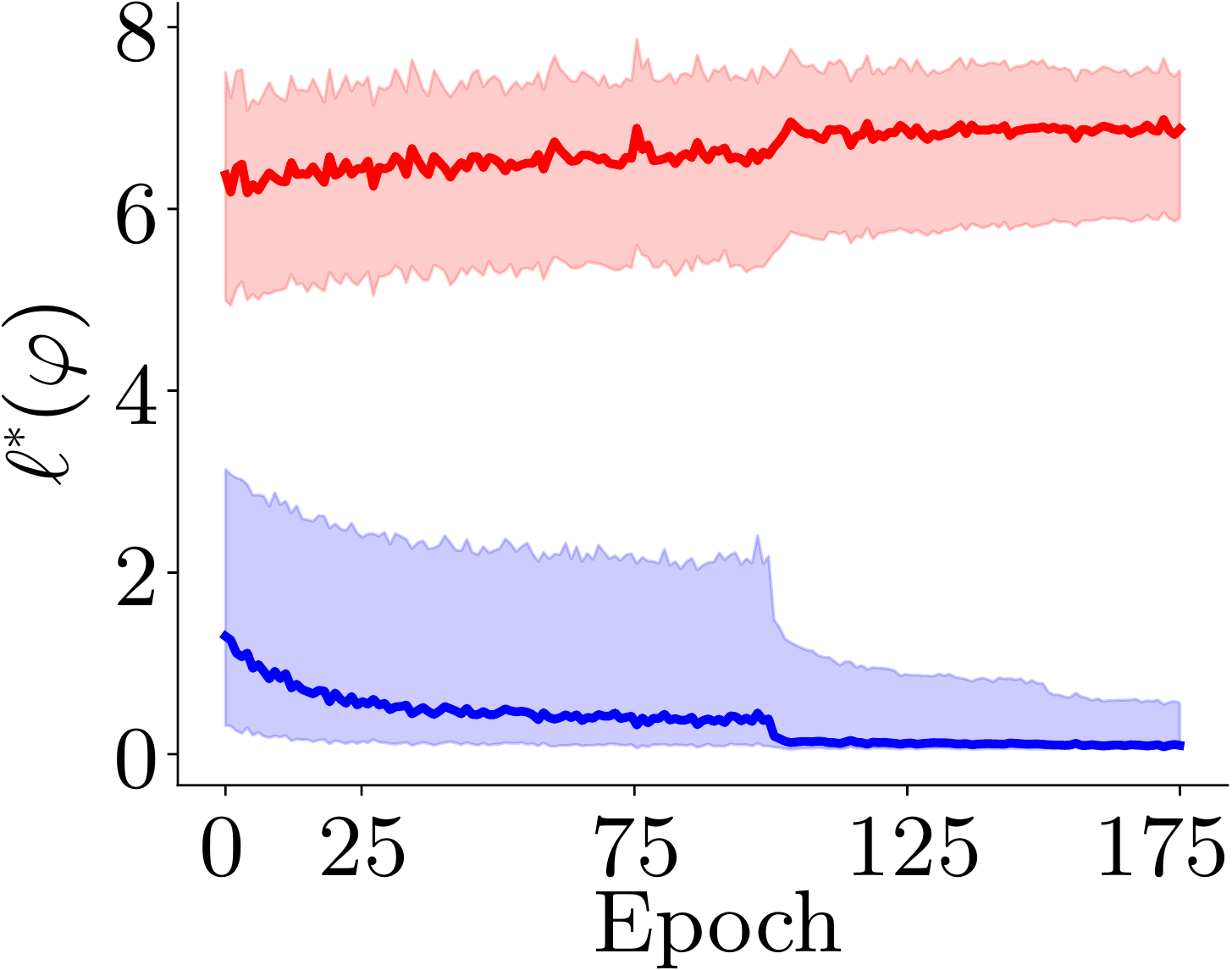}  & \includegraphics[width=0.23\textwidth]{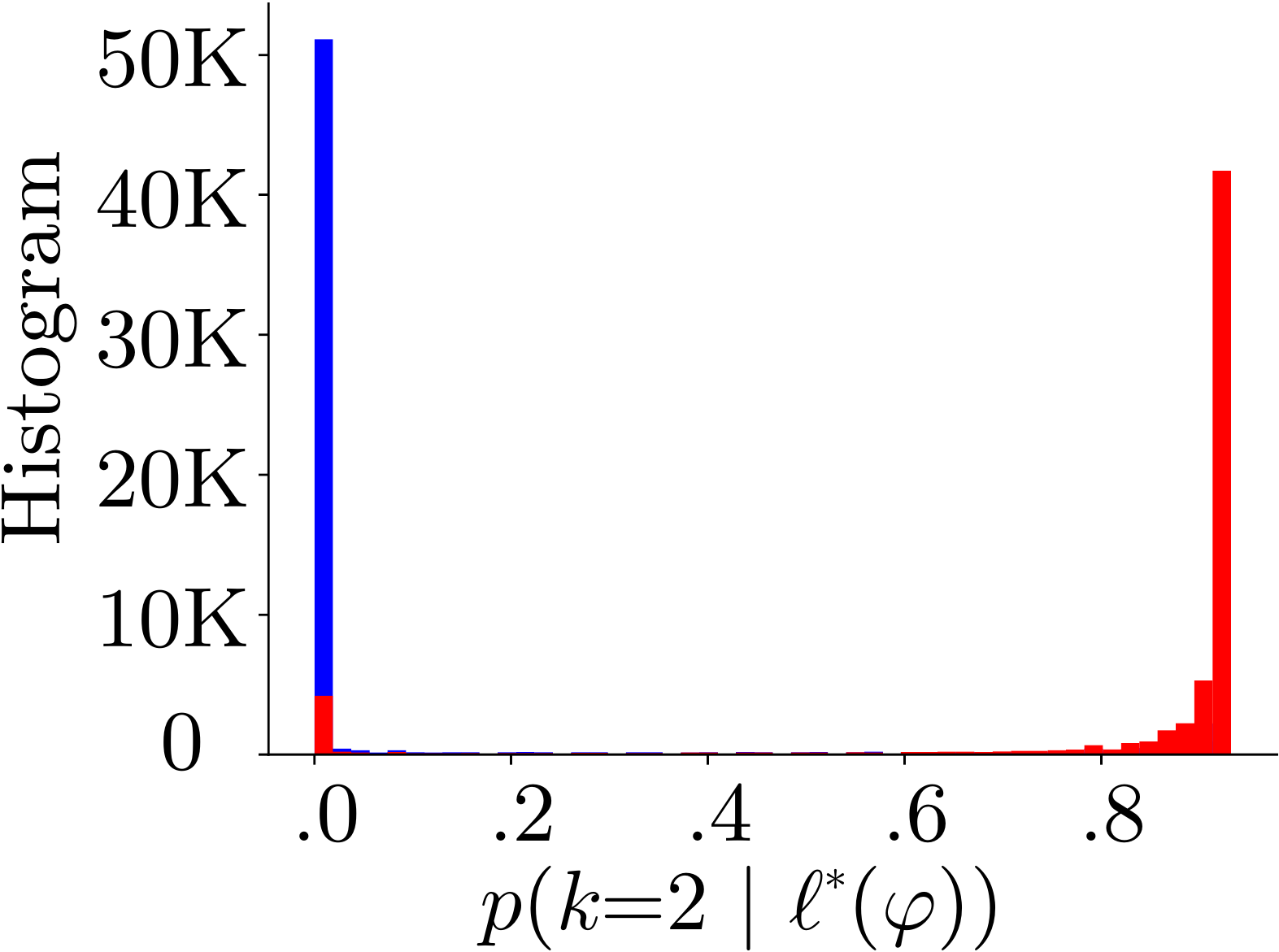}\tabularnewline
\end{tabular}\caption{\label{fig:PropoMethodLosses}Proposed approach for noise detection
in ImageNet32. Training accuracy for model $h_{\phi}(x)$ is high
due to memorization of pseudo-labels (left). Loss with respect to
the original labels $\ell^{*}\left(\phi\right)$ (mid-left) shows
that clean samples (blue) still fit the original labels $y$, while
noisy samples (red) do not. Loss $\ell^{*}\left(\varphi\right)$ for
the model $h_{\varphi}(x)$ trained in a semi-supervised manner (mid-right),
where clean/noisy samples are further separated. Posterior probability
$p\left(k=2\mid\ell^{*}\left(\varphi\right)\right)$ obtained by the
BMM (right) used to generate $\mathcal{D}_{l}$ and $\mathcal{D}_{u}$.
Figures correspond to 50\% of non-uniform in-distribution noise (see
Subsection \ref{tab:ComparisonSOTAIm32_64}).}
\end{figure*}

In \cite{2018_CVPR_JointOpt}, Eq. \eqref{eq:RelabelingLoss} is used
for robust learning against label noise and the pseudo-labels at the
end of the training are selected to start a new training stage with
improved labels. We conversely, discard noisy labels by exploiting
the observation that the new labels or pseudo-labels $\tilde{y}$
no longer represent the original label noise, but the noise from inaccurate
network predictions. This alternative view of relabeling results
in substantial improvements compared to \cite{2018_CVPR_JointOpt}
(see Subsections \ref{subsec:ImageNet32_64_eval} and \ref{subsec:CIFAR_WebEval}).
Figure \ref{fig:PropoMethodLosses} illustrates the benefits of this
approach on the detection of clean and noisy samples. The relabeling
approach progressively fits the new labels (Figure~\ref{fig:PropoMethodLosses}~(left))
as it is highly affected by confirmation bias \cite{2019_arXiv_Pseudo},
i.e. overfitting to incorrect pseudo-labels predicted by the network.
However, this reveals an interesting property in the cross-entropy
loss $\ell^{*}\left(\phi\right)=-\frac{1}{N}\sum_{i=1}^{N}y_{i}^{T}\log\left(h_{\phi}(x_{i})\right)$
with respect to the original labels $y$ (Figure \ref{fig:PropoMethodLosses}
(mid-left)) that occurs for multiple noise distributions. Pseudo-labels
(or predictions) of clean samples tend to agree with the original
labels (blue loss is low) substantially better than those of noisy
samples (red loss is high). This facilitates distinguishing between
clean and noisy samples using the loss, which does not occur in a
standard training (Figure \ref{fig:LossNoises}). Note that the training
is guided by Eq. \ref{eq:RelabelingLoss} and not by $\ell^{*}\left(\phi\right)$,
which is only used for clean/noisy discrimination.

Given the loss $\ell^{*}\left(\phi\right)$ (({*}) denotes losses
with respect the original labels $y$), we exploit the small loss
trick as in \cite{2019_ICML_BynamicBootstrapping}, i.e. we fit a
2-component Beta Mixture Model (BMM) to the loss to model clean (noisy)
samples using the first (second) component (lower losses correspond
to clean samples). The probability of each sample being clean or noisy
is then estimated using the posterior probability $p\left(k\mid\ell\right)$
from the mixture model. The initial labeled and unlabeled sets
are then defined as:{\small{}{}{}}{\small\par}

{\small{}
\begin{equation}
\mathcal{D}_{l}^{\prime}=\left\{ \left(x_{i},y_{i}\right):p\left(k=2\mid\ell_{i}^{*}\left(\phi\right)\right)\leq\gamma_{1}\right\} ,\label{eq:CleanSet}
\end{equation}
}{\small\par}

{\small{}
\begin{equation}
\mathcal{D}_{u}^{\prime}=\left\{ \left(x_{i}\right):p\left(k=2\mid\ell_{i}^{*}\left(\phi\right)\right)>\gamma_{1}\right\} ,\label{eq:NoisySet}
\end{equation}
}where $k=2$ represents noisy samples, $\ell_{i}^{*}\left(\phi\right)$
is the loss after the relabeling approach for sample $x_{i}$, and
$\gamma_{1}$ is a threshold to detect clean and noisy samples. This
threshold should be small to select only highly probable clean samples
(we use $\gamma_{1}=0.05$ unless otherwise stated). 

\subsection{Label noise detection: second stage\label{subsec:Stage2}}

The second stage of our clean-noisy samples detector refines the first
stage estimation by training a new semi-supervised model $h_{\varphi}(x)$
using the labeled and unlabeled sets from Eqs.~\eqref{eq:CleanSet}
and \eqref{eq:NoisySet}. This new model $h_{\varphi}(x)$ leads to
a loss $\ell^{*}\left(\varphi\right)$ with respect the original labels
$y$ that further facilitates the clean-noisy detection as training
is performed with far less corrupted labels. Again, we detect clean/noisy
samples by fitting a BMM to the loss $\ell^{*}\left(\varphi\right)$
and compute the final labeled and unlabeled sets $\mathcal{D}_{l}$
and $\mathcal{D}_{u}$ by selecting samples similarly to Eqs.~\eqref{eq:CleanSet}
and \eqref{eq:NoisySet}. As Figure~\ref{fig:PropoMethodLosses}
(mid-right) shows, clean and noisy samples become easily separable
in the loss $\ell^{*}\left(\varphi\right)$ leading to wider separation
in posterior probabilities $p\left(k\mid\ell^{*}\left(\varphi\right)\right)$
(Figure \ref{fig:PropoMethodLosses} (right)). This allows the use
of a higher threshold compared to $\gamma_{1}$ and lower risk of
introducing noisy samples. We therefore apply a maximum a posteriori
threshold over $p\left(k\mid\ell^{*}\left(\varphi\right)\right)$
of $\gamma_{2}=0.5$. Subsection~\ref{subsec:LabelNoiseDetection}
demonstrates the capabilities of our label noise detection method
for different noise distributions.

\subsection{The semi-supervised learning approach\label{subsec:SSLapproach}}

We adopt a recent SSL approach described in \cite{2019_arXiv_Pseudo}
for simplicity and performance. This approach performs relabeling
or pseudo-labeling as presented in the first stage, but the pseudo-labels
are only estimated for the unlabeled samples. Mixup data augmentation
\cite{2018_ICLR_mixup} is also applied to alleviate confirmation
bias and make pseudo-labeling effective. The SSL approach is applied
twice. The first time SSL is used in the second stage of the label
noise detection, i.e. with the labeled $\mathcal{D}_{l}^{\prime}$
and unlabeled $\mathcal{D}_{u}^{\prime}$ sets (see Subsection \ref{subsec:Stage1})
to train $h_{\varphi}(x)$, whereas the second time it is applied
using the final labeled and unlabeled sets $\mathcal{D}_{l}$ and
$\mathcal{D}_{u}$ (see Subsection \ref{subsec:Stage2}) to train
the final label-noise robust model $h_{\theta}(x)$.

\section{Experimental setup}

CIFAR data \cite{2009_CIFAR} is commonly corrupted for fast experimentation
with label noise, whereas real-world performance against label noise
is evaluated in datasets like WebVision \cite{2017_arXiv_WebVision}.
To the best of our knowledge, however, the related work obviates the
differences that might exist between artificially introduced noise
and real-world noise. This section introduces the proposed label noise
framework using ImageNet32/64 \cite{2009_CVPR_ImageNet} aimed at
a better understanding of label noise (Subsection \ref{subsec:ImageNet32/64})
and further describes CIFAR-10/100 and WebVision datasets as commonly
used frameworks that we also adopt (Subsections \ref{subsec:CIFAR-10/100}
and \ref{subsec:WebVision}). We use a PreAct ResNet-18 \cite{2016_ECCV_PreActResNet}
in ImageNet32/64 and CIFAR and a ResNet-18 \cite{2016_CVPR_ResNet}
for WebVision. We always train from scratch (except for the transfer
learning experiments in Subsection \ref{subsec:TransferLearning}),
using SGD with a momentum of 0.9, a weight decay of $10^{-4}$, batch
size of 128 and initial learning rate of 0.1 for every stage of our
method. Note that we do not use validation sets in any experiment.
We take this decision due to the difficulty of defining a clean validation
set in a real-world scenario and, more importantly, due to the fact
that having clean data allows direct application of SSL, which leads
to superior performance \cite{2019_arXiv_Pseudo}. See the
supplementary material\footnote{\url{https://arxiv.org/abs/1912.08741}}
for a summary of the proposed method and additional results, examples,
and configuration details.

\subsection{ImageNet32/64\label{subsec:ImageNet32/64}}

We propose to use ImageNet32/64 for fast experimentation and higher
flexibility in better understanding label noise. ImageNet32/64 are
32$\times$32 and 64$\times$64 downsampled versions of the ImageNet classification
dataset \cite{2009_CVPR_ImageNet} (1.2M images uniformly distributed
over 1000 classes). To introduce label noise we split the dataset
into $M$ in-distribution (ID) classes and $1000-M$ out-of-distribution
(OOD) classes. The split is performed to study both ID noise, as is
typically done in the literature \cite{2019_CVPR_JointOptimizImproved},
and also the less frequently considered OOD noise \cite{2018_CVPR_IterativeNoise}.
We set $M=100$ randomly selected classes that are always fixed in
our experiments, thus leading to 127K images. We study both uniform
and non-uniform noise in both ID and OOD scenarios. To introduce uniform
noise for ID we randomly flip the true label to another of the $M$
labels using uniform probabilities and excluding the true label, whereas
for OOD we randomly select a class among the $1000-M$ OOD classes
and use an image to replace the ID image. To introduce non-uniform
noise we use a label noise transition matrix \cite{2017_CVPR_ForwardLoss}
designed to be as realistic as possible. To this end, we average and
apply row-wise unit-based normalization to the confusion matrices
of the pre-trained ImageNet networks VGG-16 \cite{2014_ArXiv_VGG},
ResNet-50 \cite{2016_CVPR_ResNet}, Inception-v3 \cite{2016_CVPR_InceptionV3},
and DenseNet-161 \cite{2017_CVPR_DenseNet}. We truncate this $1000\times1000$
matrix and re-normalize it to the $M$ classes for ID noise and the
$1000-M$ classes for OOD noise. We follow the same process as the
uniform case to introduce noise, but using the row distributions corresponding
to the true label of each image: we randomly flip the label for ID
noise, while changing the image content for OOD noise. For a specific
noise level $r$, we always keep $1-r$ clean samples in each class
and modify the remainder.

We use standard data augmentation by random horizontal flips and random
4 (8) pixel translations for ImageNet32 (64) in training. During the
first stage of the label noise detection, we train for 40 epochs before
starting 60 epochs of relabeling and reduce the learning rate by a
factor of 10 in epochs 45 and 80. In the second stage of the label
noise detection, we train 175 epochs and reduce the learning rate
in epochs 100 and 150. The final SSL stage has 300 epochs with learning
rate reductions in epochs 150 and 225.

\subsection{CIFAR-10/100\label{subsec:CIFAR-10/100}}

The CIFAR-10/100 datasets \cite{2009_CIFAR} have 10/100 classes of
32$\times$32 images split into 50 (10) K images for training (testing).
We follow \cite{2019_CVPR_JointOptimizImproved} for label noise addition.
For uniform noise, labels are randomly assigned excluding the original
label. For non-uniform noise, labels are flipped with probability
$r$ to similar classes in CIFAR-10 (i.e. truck $\rightarrow$ automobile,
bird $\rightarrow$ airplane, deer $\rightarrow$ horse, cat $\rightarrow$
dog), whereas for CIFAR-100 label flips are done to the next class
circularly within the super-classes. Standard data augmentation by
random horizontal flips and random 4 pixel translations is used in
training. We train as in ImageNet32/64 in CIFAR-100, whereas in CIFAR-10
we slightly modify the first stage by training 130 epochs (70 before
relabeling) and reduce the learning rate in epochs 75 and 110. This
increase in training epochs is to ensure a better model before relabeling,
as classes in CIFAR-10 are more different than in ImageNet32/64 and
CIFAR-100, thus incorrect predictions during relabeling are less informative.
We use $\gamma_{1}=0.1$ to assure sufficient labeled samples for
SSL.

\subsection{WebVision\label{subsec:WebVision}}

We use WebVision 1.0 \cite{2017_arXiv_WebVision} to evaluate performance
on real-world label noise. We evaluate our approach using a subset
of classes as done in other works \cite{2019_ICML_WebVision50,2020_ICLR_DivideMix}.
We select all WebVision data for the first 50 classes resulting in
a dataset of 137K images. We use random horizontal flips during training
and resize images to $256\times256$ before taking random $224\times224$
crops. For the first stage of the label noise detection, we train
40 epochs before starting 60 epochs of relabeling and reduce the learning
rate dividing by 10 twice (epochs 45 and 80). For the second stage,
we train 150 epochs and reduce the learning rate in epochs 100 and
125. The final SSL stage has 200 epochs with learning rate reductions
in epochs 150 and 175.

\section{Experiments}

\subsection{Label noise detection comparison\label{subsec:LabelNoiseDetection}}

\begin{figure*}[t]
\centering{}\setlength{\tabcolsep}{0.0pt} 
\global\long\def\arraystretch{1}%
\resizebox{1.0\textwidth}{!}{%
\begin{tabular}{cccc}
\includegraphics[width=0.25\textwidth]{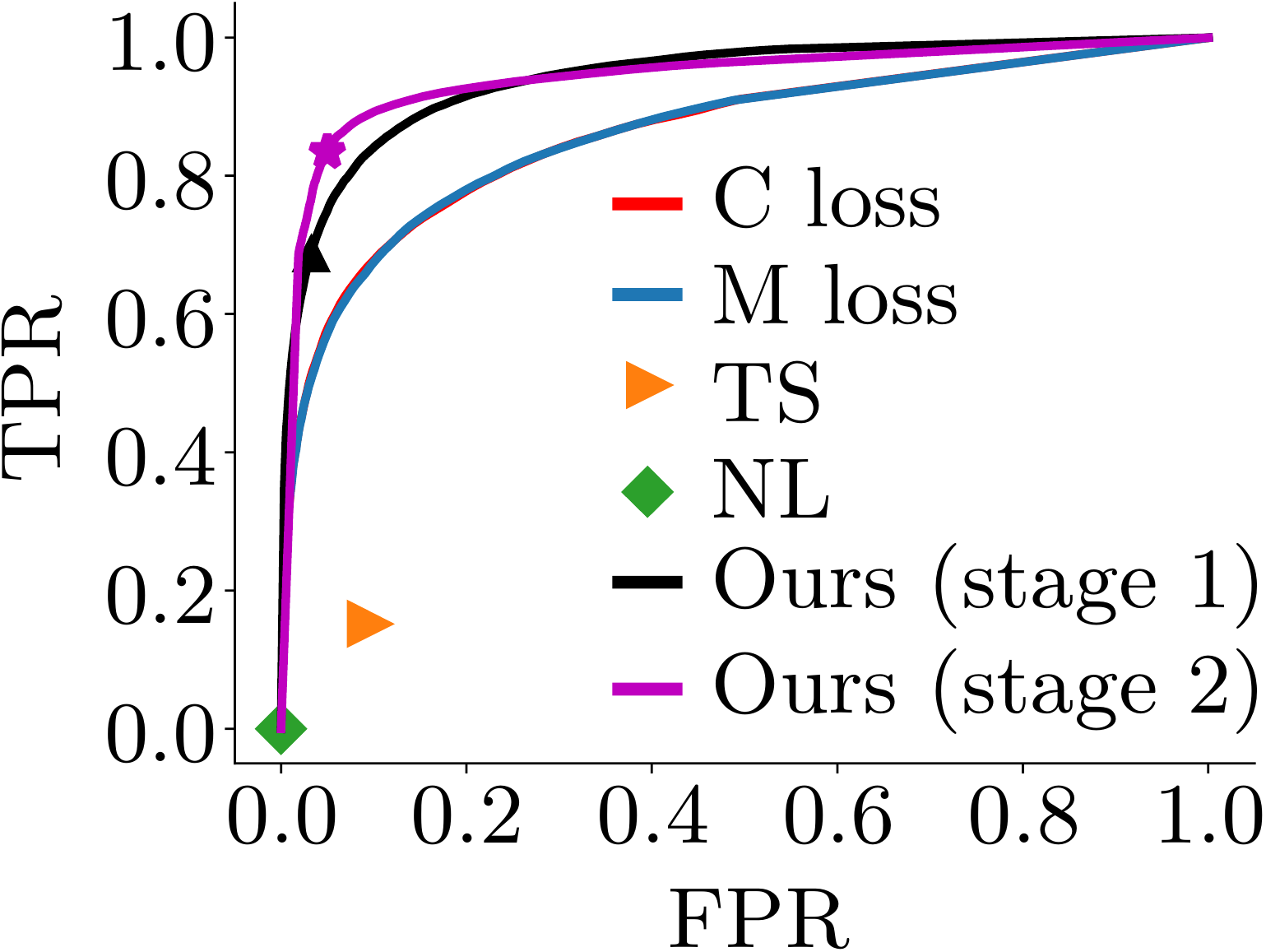} & \includegraphics[width=0.25\textwidth]{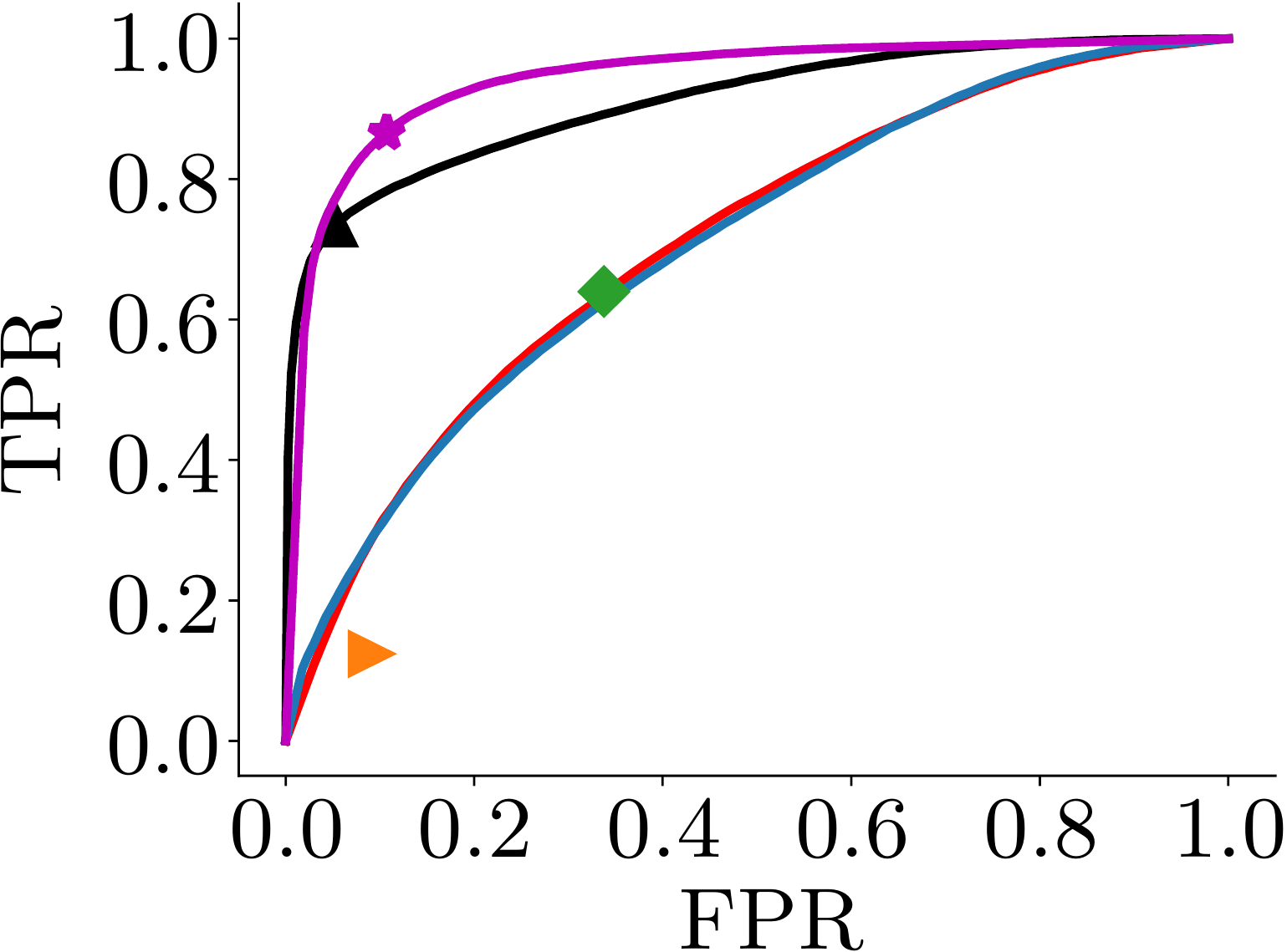} & \includegraphics[width=0.25\textwidth]{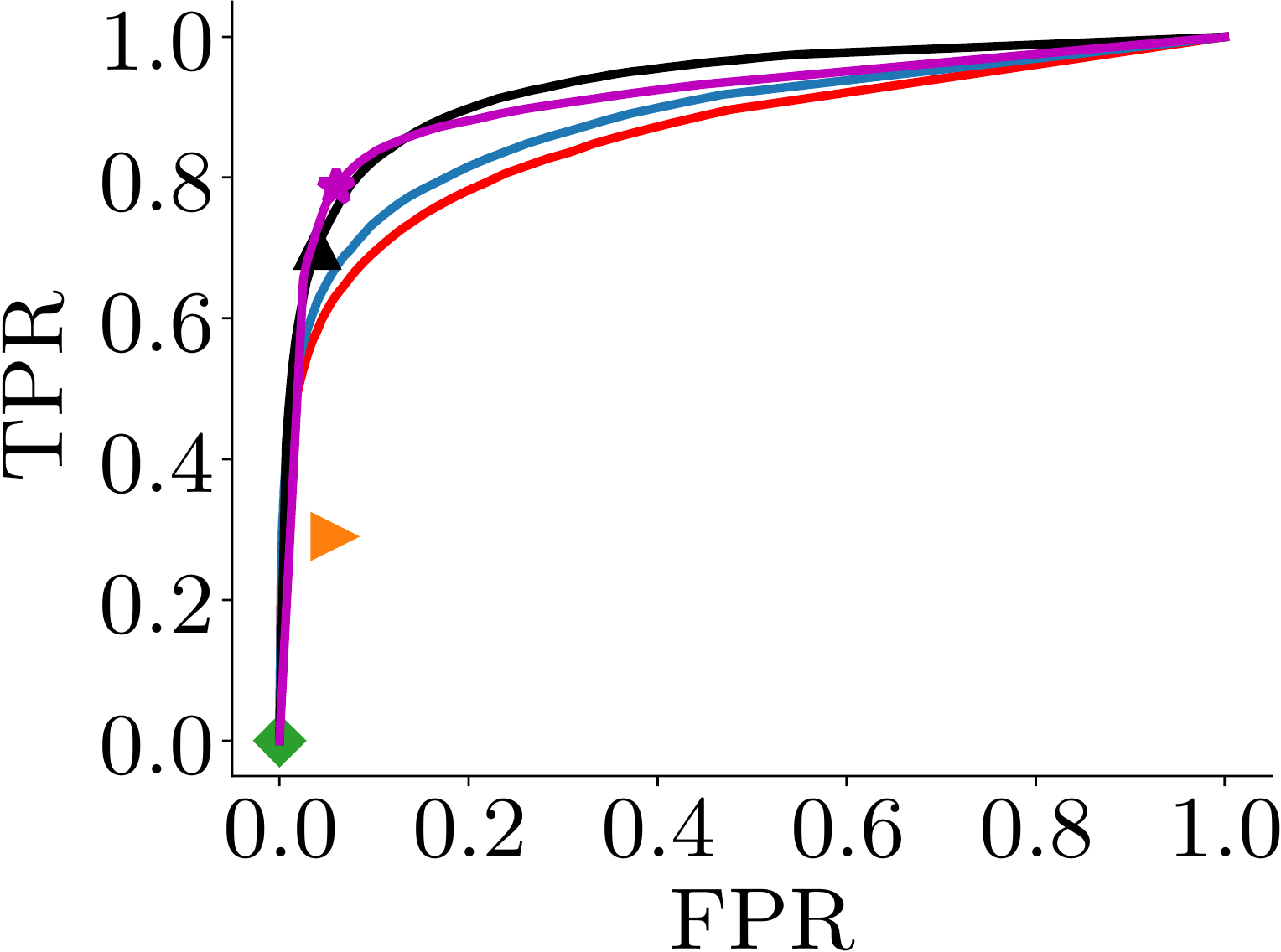} & \includegraphics[width=0.25\textwidth]{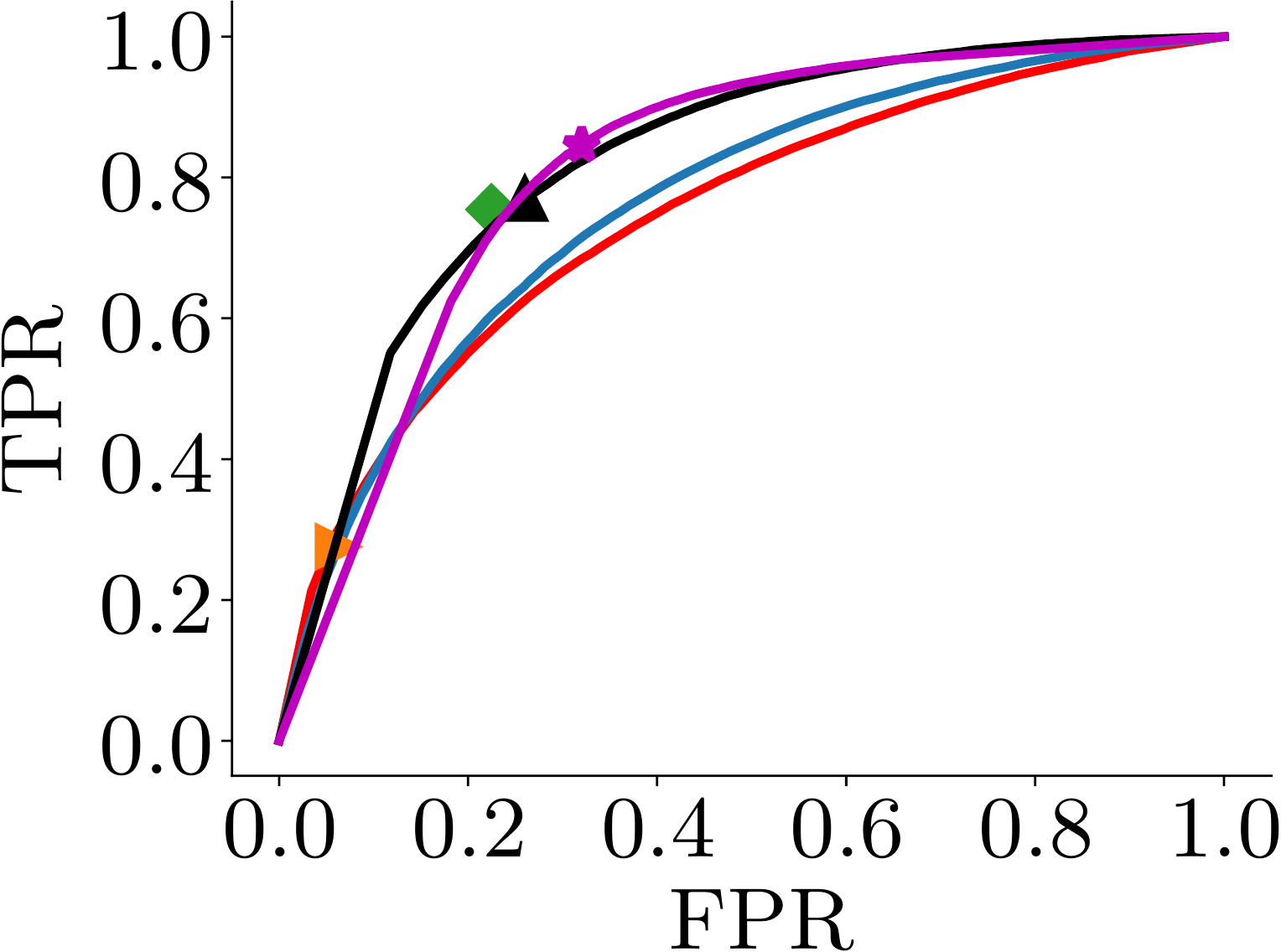}\tabularnewline
\end{tabular}}\caption{\label{fig:Label-noise-detection}Label noise detection methods in
ImageNet32. 80\% uniform ID (left) and OOD (mid-left) noise. 50\%
non-uniform ID (mid-right) and OOD (right) noise. The proposed method
(star in stage 2 denotes operating point) surpasses detection based
on cross-entropy loss without (C) and with mixup (M), TS \cite{2018_WACV_SemiSupNoise}
and NL \cite{2019_ICCV_NegativeLearning}. Key: ID (in-distribution).
OOD (out-of-distribution).}
\end{figure*}
\begin{table}[t]
\begin{centering}
\caption{\label{tab:LabelNoiseDetection}Label noise detection results in ImageNet32.
Key: NU (non-uniform); U (uniform); ID (in-distribution); OOD (out-of-distribution).}
\par\end{centering}
\noindent \begin{centering}
{\footnotesize{}\centering}
\global\long\def\arraystretch{0.75}%
{\footnotesize{}\setlength{\tabcolsep}{1.0pt}}\resizebox{0.90\columnwidth}{!}{{\footnotesize{}}%
\begin{tabular}{llccccccc}
\toprule 
 &  & \multicolumn{3}{c}{{\footnotesize{}NU-ID}} & \multicolumn{4}{c}{{\footnotesize{}U-ID}}\tabularnewline
\midrule 
 &  & {\footnotesize{}10\%} & {\footnotesize{}30\%} & {\footnotesize{}50\%} & {\footnotesize{}20\%} & {\footnotesize{}40\%} & {\footnotesize{}60\%} & {\footnotesize{}80\%}\tabularnewline
\midrule 
\multirow{2}{*}{{\footnotesize{}TS \cite{2018_WACV_SemiSupNoise}}} & {\footnotesize{}TPR} & {\footnotesize{}0.28} & {\footnotesize{}0.14} & {\footnotesize{}0.12} & {\footnotesize{}0.23} & {\footnotesize{}0.14} & {\footnotesize{}0.15} & {\footnotesize{}0.15}\tabularnewline
 & {\footnotesize{}FPR} & {\footnotesize{}0.04} & {\footnotesize{}0.07} & {\footnotesize{}0.09} & {\footnotesize{}0.08} & {\footnotesize{}0.06} & {\footnotesize{}0.07} & {\footnotesize{}0.09}\tabularnewline
\multirow{2}{*}{{\footnotesize{}NL \cite{2019_ICCV_NegativeLearning}}} & {\footnotesize{}TPR} & {\footnotesize{}0.80} & {\footnotesize{}0.75} & {\footnotesize{}0.64} & {\footnotesize{}0.83} & {\footnotesize{}0.77} & {\footnotesize{}0.34} & {\footnotesize{}-}\tabularnewline
 & {\footnotesize{}FPR} & {\footnotesize{}0.09} & {\footnotesize{}0.26} & {\footnotesize{}0.34} & {\footnotesize{}0.03} & \textbf{\footnotesize{}0.03} & \textbf{\footnotesize{}0.00} & {\footnotesize{}-}\tabularnewline
\midrule 
\multirow{2}{*}{{\footnotesize{}Ours}} & {\footnotesize{}TPR} & \textbf{\footnotesize{}0.82} & \textbf{\footnotesize{}0.90} & \textbf{\footnotesize{}0.87} & \textbf{\footnotesize{}0.89} & \textbf{\footnotesize{}0.94} & \textbf{\footnotesize{}0.92} & \textbf{\footnotesize{}0.83}\tabularnewline
 & {\footnotesize{}FPR} & \textbf{\footnotesize{}0.02} & \textbf{\footnotesize{}0.04} & \textbf{\footnotesize{}0.11} & \textbf{\footnotesize{}0.02} & {\footnotesize{}0.05} & {\footnotesize{}0.05} & {\footnotesize{}0.05}\tabularnewline
\bottomrule
\end{tabular}}
\par\end{centering}
\medskip{}

\noindent \centering{}{\footnotesize{}\centering}
\global\long\def\arraystretch{0.75}%
{\footnotesize{}\setlength{\tabcolsep}{1.0pt}}\resizebox{0.90\columnwidth}{!}{{\footnotesize{}}%
\begin{tabular}{llccccccc}
\toprule 
 &  & \multicolumn{3}{c}{{\footnotesize{}NU-OOD}} & \multicolumn{4}{c}{{\footnotesize{}U-OOD}}\tabularnewline
\midrule 
 &  & {\footnotesize{}10\%} & {\footnotesize{}30\%} & {\footnotesize{}50\%} & {\footnotesize{}20\%} & {\footnotesize{}40\%} & {\footnotesize{}60\%} & {\footnotesize{}80\%}\tabularnewline
\midrule 
\multirow{2}{*}{{\footnotesize{}TS \cite{2018_WACV_SemiSupNoise}}} & {\footnotesize{}TPR} & {\footnotesize{}0.40} & {\footnotesize{}0.36} & {\footnotesize{}0.27} & {\footnotesize{}0.38} & {\footnotesize{}0.32} & {\footnotesize{}0.30} & {\footnotesize{}0.29}\tabularnewline
 & {\footnotesize{}FPR} & \textbf{\footnotesize{}0.04} & \textbf{\footnotesize{}0.06} & \textbf{\footnotesize{}0.06} & \textbf{\footnotesize{}0.03} & \textbf{\footnotesize{}0.02} & {\footnotesize{}0.02} & \textbf{\footnotesize{}0.06}\tabularnewline
\multirow{2}{*}{{\footnotesize{}NL \cite{2019_ICCV_NegativeLearning}}} & {\footnotesize{}TPR} & {\footnotesize{}0.82} & {\footnotesize{}0.81} & {\footnotesize{}0.75} & {\footnotesize{}0.80} & {\footnotesize{}0.79} & {\footnotesize{}0.66} & {\footnotesize{}-}\tabularnewline
 & {\footnotesize{}FPR} & {\footnotesize{}0.10} & {\footnotesize{}0.18} & {\footnotesize{}0.22} & \textbf{\footnotesize{}0.03} & {\footnotesize{}0.03} & \textbf{\footnotesize{}0.01} & {\footnotesize{}-}\tabularnewline
\midrule 
\multirow{2}{*}{{\footnotesize{}Ours}} & {\footnotesize{}TPR} & \textbf{\footnotesize{}0.83} & \textbf{\footnotesize{}0.86} & \textbf{\footnotesize{}0.85} & \textbf{\footnotesize{}0.89} & \textbf{\footnotesize{}0.90} & \textbf{\footnotesize{}0.87} & \textbf{\footnotesize{}0.79}\tabularnewline
 & {\footnotesize{}FPR} & {\footnotesize{}0.07} & {\footnotesize{}0.23} & {\footnotesize{}0.32} & {\footnotesize{}0.04} & {\footnotesize{}0.06} & {\footnotesize{}0.06} & \textbf{\footnotesize{}0.06}\tabularnewline
\bottomrule
\end{tabular}}
\end{table}
Transforming the supervised training with label noise into SSL requires
detecting the noise to, ideally, discard the labels and turn noisy
samples into unlabeled ones. As commented in Section \ref{sec:Introduction},
many approaches use the small loss trick, i.e. considering low loss
samples as clean ones, to accomplish such detection. However, Figure
\ref{fig:LossNoises} shows that different noise distributions present
different challenges, limiting a straightforward application of this
trick. We confirm this limitation in Figure \ref{fig:Label-noise-detection},
where we compare, for different label noise distributions, the Receiver
Operating Characteristic (ROC) curves for our label noise detection
method and the small loss trick (using cross-entropy with and without
mixup \cite{2018_ICLR_mixup}). A straightforward application of the
small loss trick \cite{2019_ICML_BynamicBootstrapping,2018_NeurIPS_CoTeaching,2019_ICML_SELFIE}
(C and M losses) is relatively robust for uniform noises (left and
mid-right), but poorly performs for non-uniform ones (mid-left and
right). Conversely, the proposed approach robustly detects noisy samples
for different noise distributions, thus supporting the effectiveness
of our method. We also outperform two recently proposed label noise
detection methods \cite{2018_WACV_SemiSupNoise,2019_ICCV_NegativeLearning}
across different label noise distributions (see Table \ref{tab:LabelNoiseDetection}).
Note that we encounter some limitations addressing high levels of
non-uniform OOD noise, which occurs due to the nature of the classes
used as noise. We are using the ImageNet confusion matrix in the validation
set to introduce noise and we have 100 (900) in- (out-of-) distribution
classes, thus using the most challenging classes as OOD noise.

\subsection{Comparison with related work\label{subsec:ImageNet32_64_eval}}

\begin{figure}[t]
\centering{}\includegraphics[width=1\columnwidth]{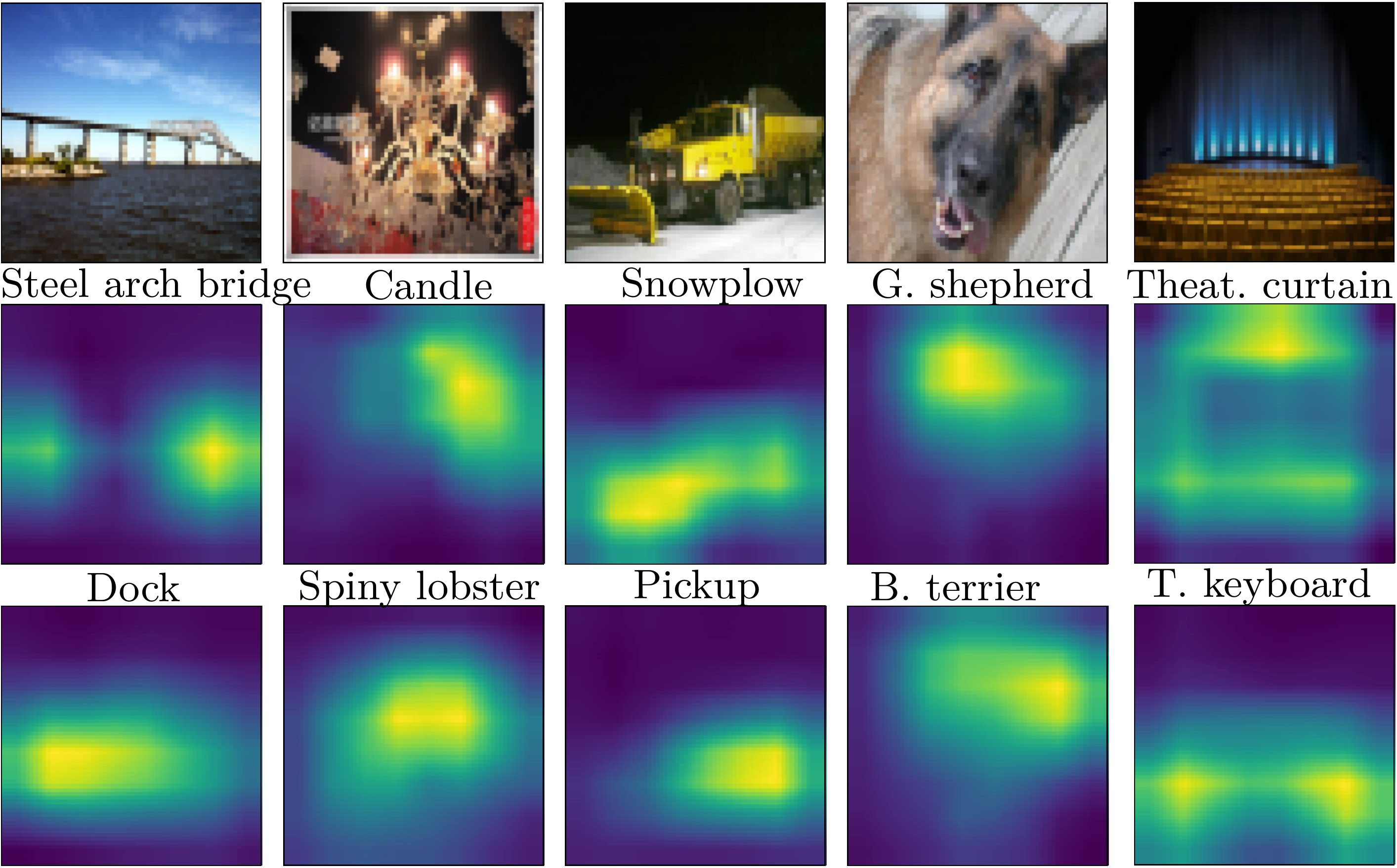}\caption{\label{fig:VisualizationCAM}Label noise memorization for undetected
noisy images in ImageNet64 (50\%, non-uniform in-distribution noise).
From top to bottom: image and Class Activation Map (CAM) for the true
and predicted class. Note that the predicted class (bottom) is also
the noisy label used during training.}
\end{figure}
{\small{}}
\begin{table*}[t]
\begin{centering}
{\small{}\caption{\label{tab:ComparisonSOTAIm32_64}ImageNet32/64 accuracy. Bold denotes
best performance. Key: NU (non-uniform); U (uniform); ID (in-distribution);
OOD (out-of-distribution).}
}{\small\par}
\par\end{centering}
{\small{}\centering\setlength{\tabcolsep}{0.1pt}\renewcommand*{\arraystretch}{1.0}}\resizebox{1\textwidth}{!}{

{\footnotesize{}}%
\begin{tabular}{lcccccccccccccccc}
\toprule 
 & \multicolumn{8}{c|}{{\footnotesize{}ImageNet32}} & \multicolumn{8}{c}{{\footnotesize{}ImageNet64}}\tabularnewline
\midrule 
 & \multicolumn{2}{c}{{\footnotesize{}NU-ID}} & \multicolumn{2}{c}{{\footnotesize{}U-ID}} & \multicolumn{2}{c}{{\footnotesize{}NU-OOD}} & \multicolumn{2}{c}{{\footnotesize{}U-OOD}} & \multicolumn{2}{c}{{\footnotesize{}NU-ID}} & \multicolumn{2}{c}{{\footnotesize{}U-ID}} & \multicolumn{2}{c}{{\footnotesize{}NU-OOD}} & \multicolumn{2}{c}{{\footnotesize{}U-OOD}}\tabularnewline
\midrule 
 & {\footnotesize{}30\%} & {\footnotesize{}50\%} & {\footnotesize{}40\%} & {\footnotesize{}80\%} & {\footnotesize{}30\%} & {\footnotesize{}50\%} & {\footnotesize{}40\%} & {\footnotesize{}80\%} & {\footnotesize{}30\%} & {\footnotesize{}50\%} & {\footnotesize{}40\%} & {\footnotesize{}80\%} & {\footnotesize{}30\%} & {\footnotesize{}50\%} & {\footnotesize{}40\%} & {\footnotesize{}80\%}\tabularnewline
\midrule 
{\footnotesize{}FW \cite{2017_CVPR_ForwardLoss} } & {\footnotesize{}54.22 } & {\footnotesize{}43.38 } & {\footnotesize{}52.06 } & {\footnotesize{}31.20 } & {\footnotesize{}62.14 } & {\footnotesize{}55.06 } & {\footnotesize{}56.32 } & {\footnotesize{}40.08 } & {\footnotesize{}60.10} & {\footnotesize{}46.06} & {\footnotesize{}57.42} & {\footnotesize{}37.84} & {\footnotesize{}69.86} & {\footnotesize{}63.38} & {\footnotesize{}63.08} & {\footnotesize{}47.68}\tabularnewline
{\footnotesize{}R \cite{2018_CVPR_JointOpt} } & {\footnotesize{}67.24 } & {\footnotesize{}63.62 } & {\footnotesize{}62.98 } & {\footnotesize{}41.52 } & {\footnotesize{}66.36 } & {\footnotesize{}62.80 } & {\footnotesize{}64.04 } & {\footnotesize{}45.00 } & {\footnotesize{}74.28} & {\footnotesize{}69.20} & {\footnotesize{}70.98} & {\footnotesize{}48.44} & {\footnotesize{}74.22} & {\footnotesize{}70.74} & {\footnotesize{}72.78} & {\footnotesize{}54.00}\tabularnewline
{\footnotesize{}M \cite{2018_ICLR_mixup} } & {\footnotesize{}67.14 } & {\footnotesize{}51.96 } & {\footnotesize{}61.98 } & {\footnotesize{}38.92 } & {\footnotesize{}66.14 } & {\footnotesize{}60.62 } & {\footnotesize{}64.66 } & {\footnotesize{}47.40 } & {\footnotesize{}74.02} & {\footnotesize{}58.14} & {\footnotesize{}69.90} & {\footnotesize{}49.22} & {\footnotesize{}74.78} & {\footnotesize{}69.40} & {\footnotesize{}73.94} & {\footnotesize{}59.54}\tabularnewline
{\footnotesize{}DB \cite{2019_ICML_BynamicBootstrapping} } & {\footnotesize{}62.88 } & {\footnotesize{}52.20 } & {\footnotesize{}67.62 } & {\footnotesize{}45.34 } & {\footnotesize{}64.86 } & {\footnotesize{}60.58 } & {\footnotesize{}65.96 } & {\footnotesize{}39.30 } & {\footnotesize{}71.30} & {\footnotesize{}60.98} & {\footnotesize{}74.56} & {\footnotesize{}56.44} & {\footnotesize{}77.94} & {\footnotesize{}70.38} & {\footnotesize{}74.08} & {\footnotesize{}50.98}\tabularnewline
\midrule 
{\footnotesize{}DRPL } & \textbf{\footnotesize{}73.46}{\footnotesize{} } & \textbf{\footnotesize{}68.18}{\footnotesize{} } & \textbf{\footnotesize{}73.48}{\footnotesize{} } & \textbf{\footnotesize{}61.78}{\footnotesize{} } & \textbf{\footnotesize{}71.38}{\footnotesize{} } & \textbf{\footnotesize{}67.32}{\footnotesize{} } & \textbf{\footnotesize{}71.36}{\footnotesize{} } & \textbf{\footnotesize{}54.10}{\footnotesize{} } & \textbf{\footnotesize{}81.90} & \textbf{\footnotesize{}77.66} & \textbf{\footnotesize{}81.50} & \textbf{\footnotesize{}73.08} & \textbf{\footnotesize{}80.44} & \textbf{\footnotesize{}76.38} & \textbf{\footnotesize{}79.76} & \textbf{\footnotesize{}64.34}\tabularnewline
\bottomrule 
\end{tabular}}
\end{table*}
{\small\par}

We select representative top-performing loss correction \cite{2019_ICML_BynamicBootstrapping,2017_CVPR_ForwardLoss},
relabeling \cite{2018_CVPR_JointOpt}, and label noise robust regularization
approaches \cite{2018_ICLR_mixup} to compare against our label noise
Distribution Robust Pseudo-Labeling (DRPL) approach in Table \ref{tab:ComparisonSOTAIm32_64}.
The proposed method gives remarkable improvements across different
levels and distributions of label noise. Note that, unlike most methods
compared, our method shows little degradation between the best (reported
in Table \ref{tab:ComparisonSOTAIm32_64}) and last epoch accuracy
(see extended results in the supplementary material). In general,
R \cite{2018_CVPR_JointOpt} behaves consistently across label noise
levels and distributions, while DB \cite{2019_ICML_BynamicBootstrapping}
has problems with non-uniform noise. FW \cite{2017_CVPR_ForwardLoss}
and M \cite{2018_ICLR_mixup} tend to exhibit worse performance. An
important observation is that non-uniform OOD noise exhibits less degradation than other noise types for all methods. This is reasonable
as OOD samples whose content is close to an ID class will contribute
to improved representation learning due to semantic similarities,
and the network predictions for these samples will not harm the model
for ID classes. This behavior resembles real-world noisy data as observed
in the WebVision dataset results in Subsection \ref{subsec:CIFAR_WebEval}.

\subsection{Effects of label noise memorization\label{subsec:TransferLearning}}

We know that clean (easier) patterns are learned first by deep neural
networks and that noisy labels can be completely memorized \cite{2017_ICML_Memorization,2017_ICLR_Rethinking}.
\emph{How do networks memorize noisy samples?} Figure~\ref{fig:VisualizationCAM}
provides some intuition. It shows the Class Activation Maps \cite{2016_CVPR_CAMs}
for the true class and the predicted class for undetected noisy samples.
The network skips relevant areas for the true class, attending to
areas that might help explain (i.e. memorizing) the noisy label. For
example, the network extends the class activation maps to cover larger
regions (columns 2, 4) such as the whole lamp instead of the candles
for the noisy label \emph{lobster} (column 2), while other times it
also omits characteristic areas of the true class (columns 1, 3 and
5) such as the right part of the bridge to better fit the noisy label
\emph{dock} (column 1).

\begin{figure*}[t]
\centering{}\setlength{\tabcolsep}{1pt}%
\begin{tabular}{cccc}
\includegraphics[width=0.24\textwidth]{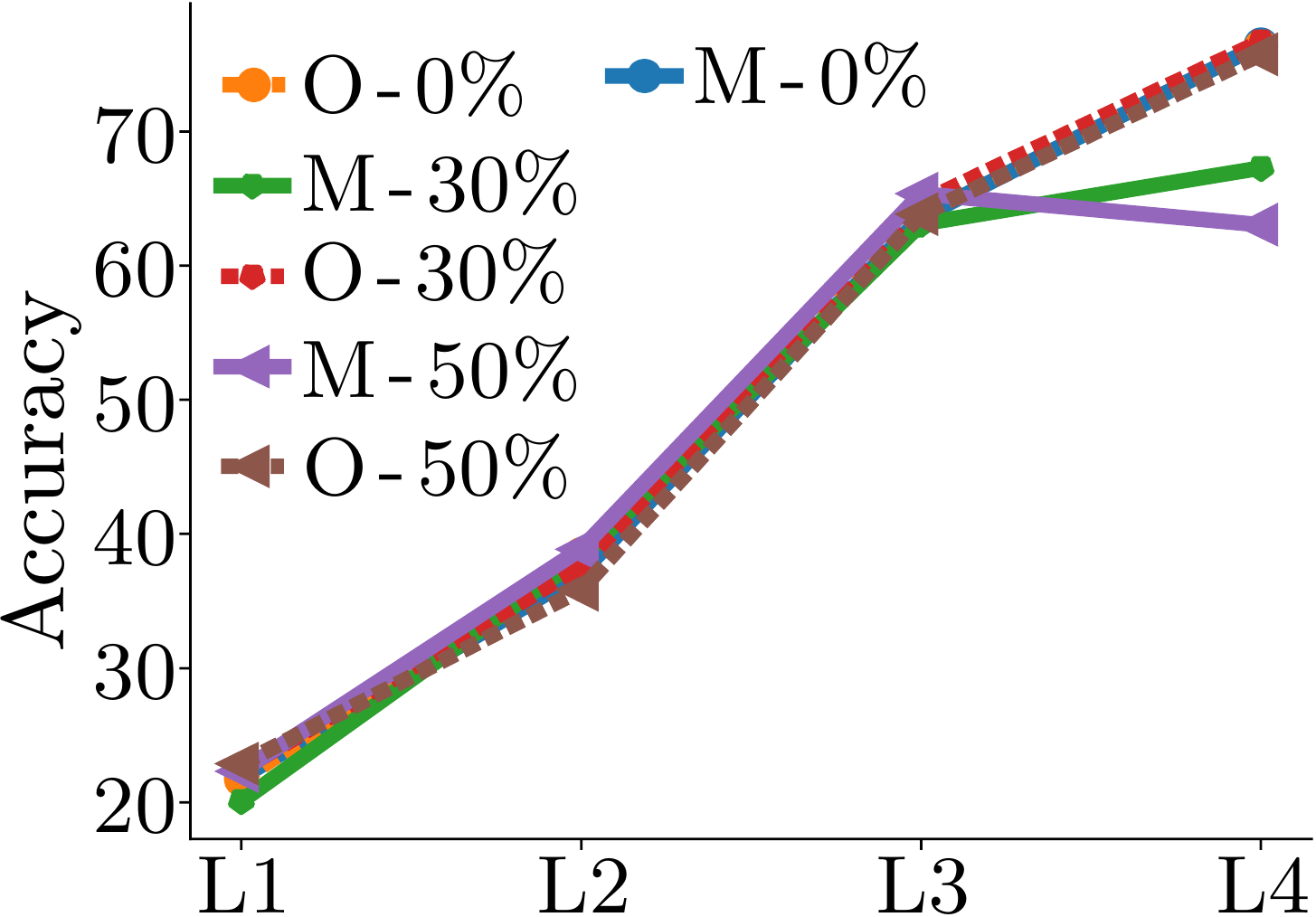}  & \includegraphics[width=0.24\textwidth]{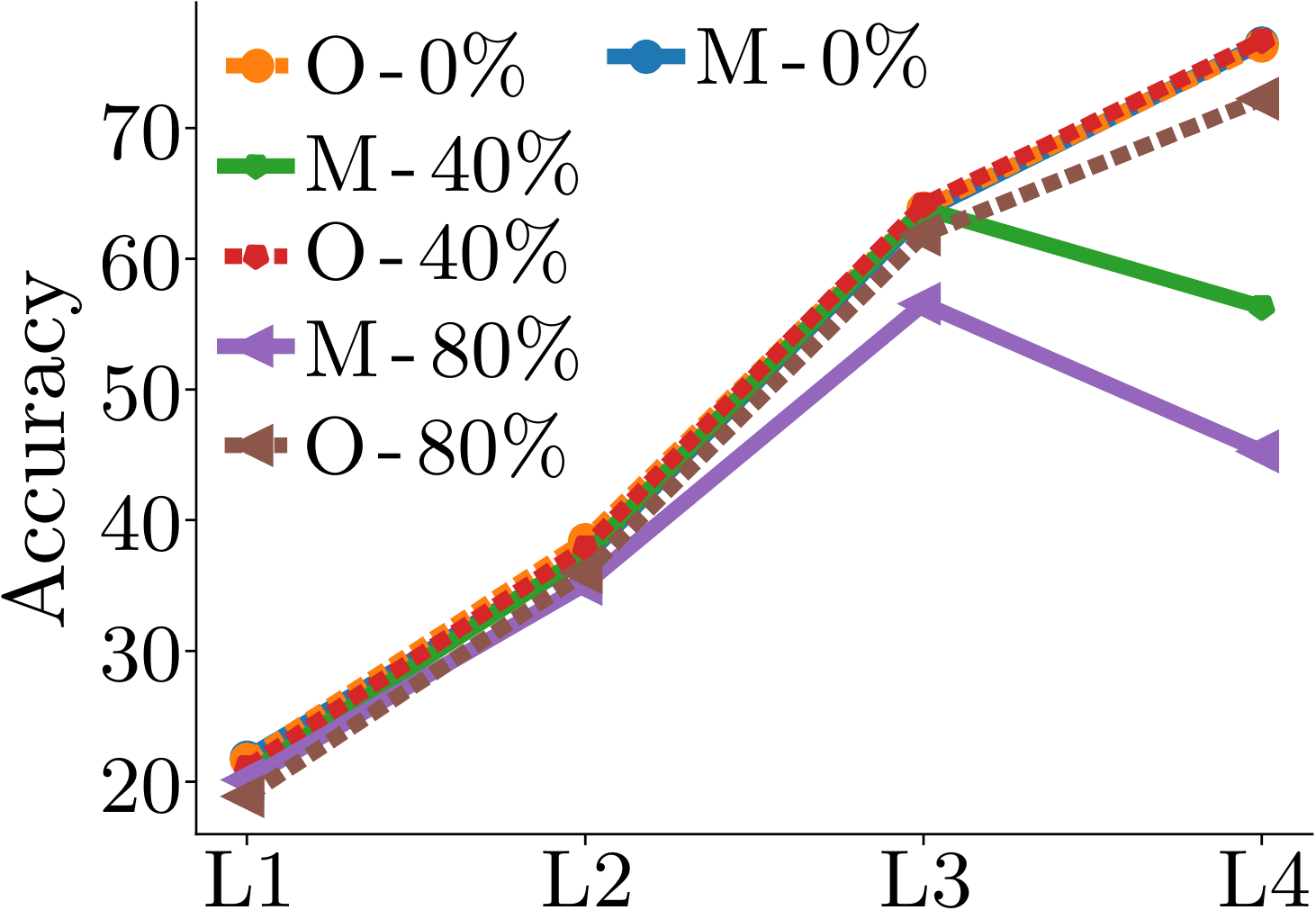}  & \includegraphics[width=0.24\textwidth]{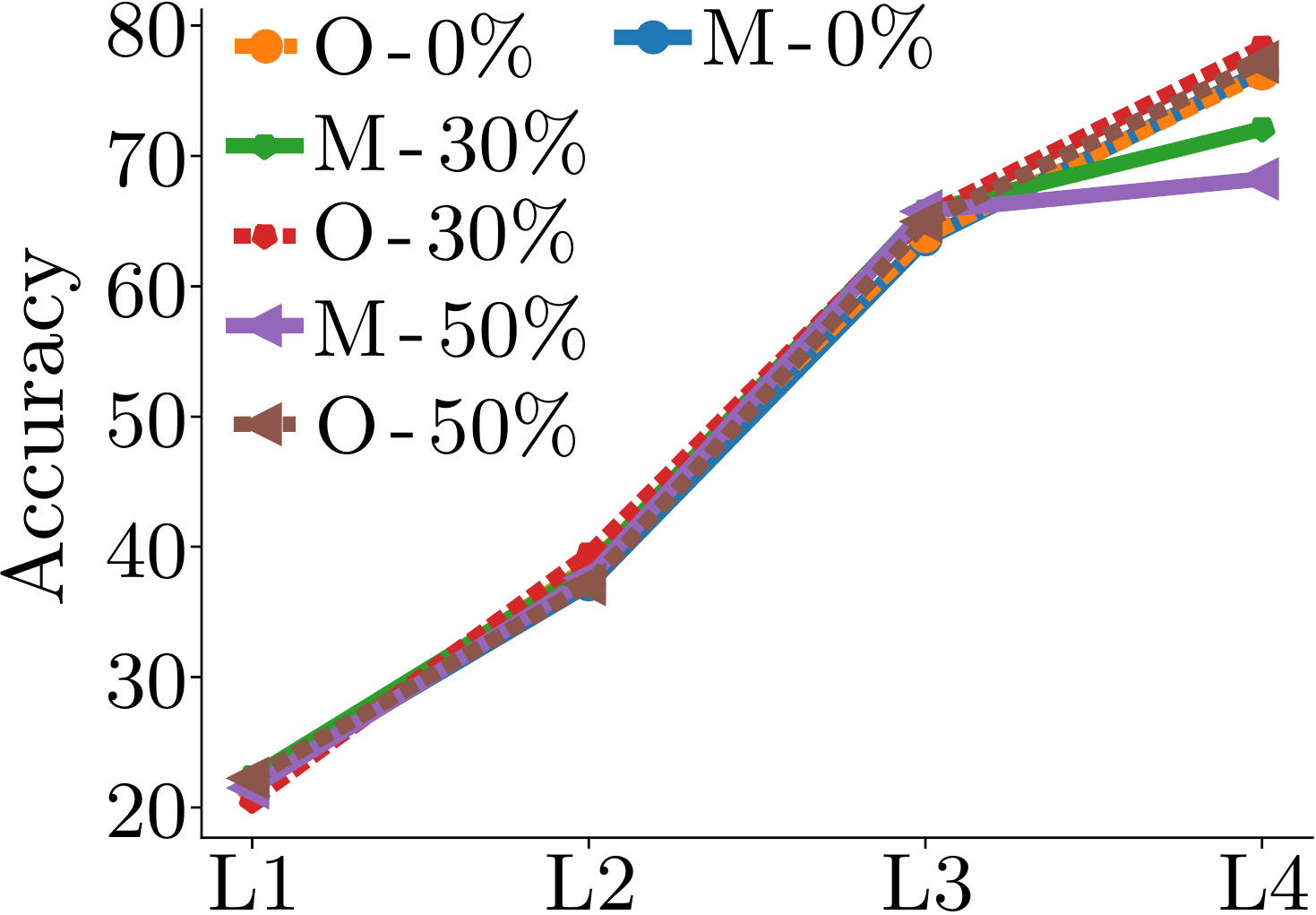}  & \includegraphics[width=0.24\textwidth]{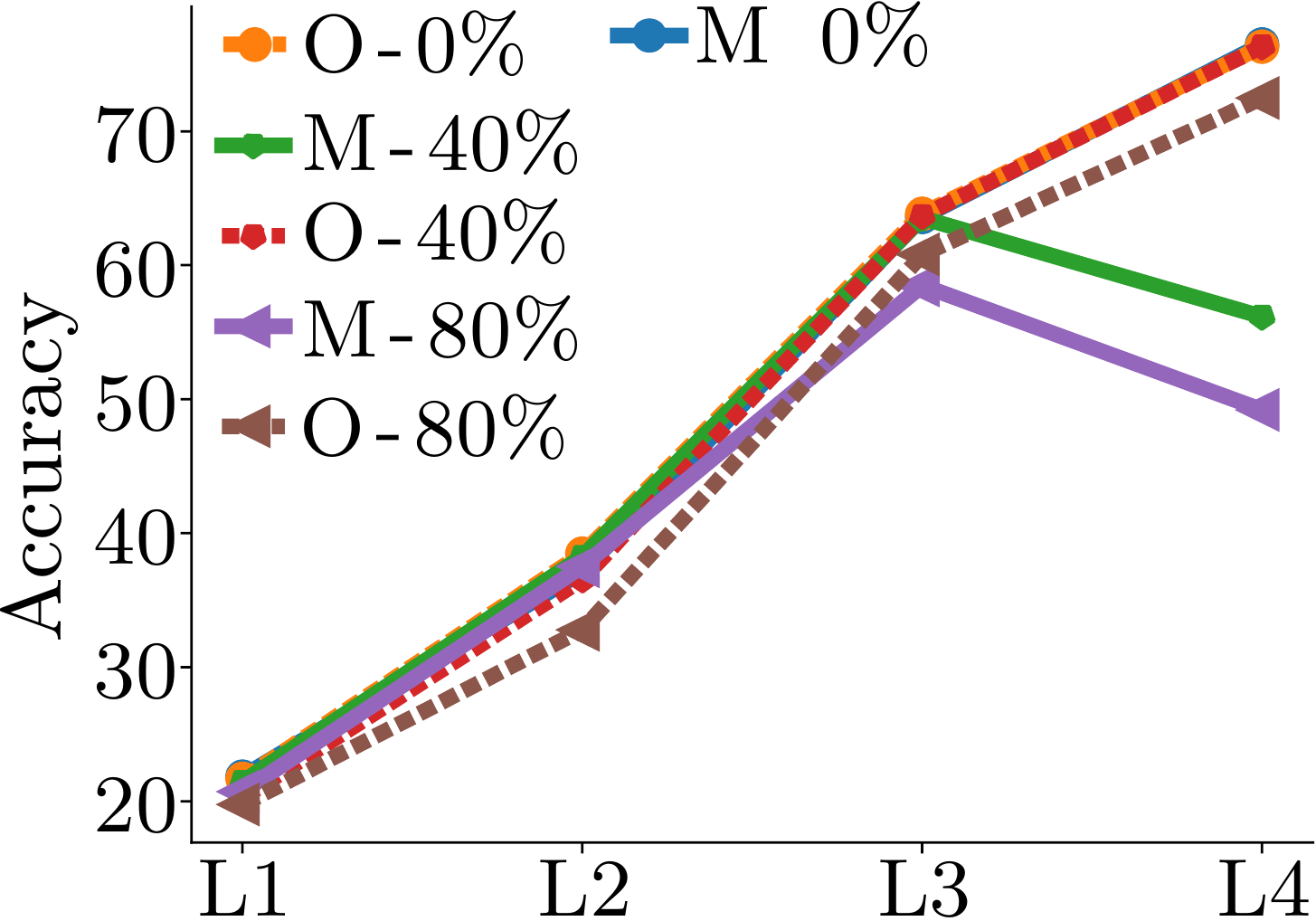}\tabularnewline
\end{tabular}\caption{\label{fig:TransferFigures} ImageNet64 linear probes. A linear classifier
is trained using features at different depths. Source task: 100 classes.
Target task: 25 classes from the remaining 900. Noise: in-distribution
non-uniform (left) and uniform (mid-left), and out-of-distribution
non-uniform (mid-right) and uniform (right). Model from the last training
epoch in the source domain is used (i.e. M has fitted the noise).
Key: M: Mixup. O: Ours.}
\end{figure*}

\emph{Does memorization negatively impact visual representation learning?}
We follow the standard approach of using linear probes \cite{2017_CVPR_LinearProbes}
to verify the utility of features under different noise levels and
distributions in a target task. Specifically, we train a linear classifier
on the global average pooled activations obtained after each of the
4 PreAct ResNet-18 blocks. Figure~\ref{fig:TransferFigures} presents
the results in ImageNet64. Our model (O) clearly outperforms mixup
(M) in the target task performance using features from the last block.
Better source models (i.e. trained with less noise) also tend to produce
better target performance. However, an interesting finding is that
for both uniform and non-uniform noise, the final accuracy exhibits
degradation in L4, while for earlier features no degradation is observed
even for M (a model that has memorized the noise). An exception is
80\% of uniform noise, where degradation is found across all blocks,
but with more discriminative representations learned by our approach.
Similar results are observed in ImageNet32 (see supplementary material).
\begin{table*}[t]
\begin{centering}
\caption{\label{tab:WebV}WebVision (50 classes) accuracy. Key: NU (non-uniform
noise). U (uniform noise). CE (Cross-entropy).}
\par\end{centering}
\centering{}\setlength{\tabcolsep}{3.0pt} \resizebox{0.66\textwidth}{!}{{\footnotesize{}}%
\begin{tabular}{rrrrrrrrrr}
\toprule 
{\footnotesize{}CE} & {\footnotesize{}FW \cite{2017_CVPR_ForwardLoss}} & {\footnotesize{}R \cite{2018_CVPR_JointOpt}} & {\footnotesize{}M \cite{2018_ICLR_mixup}} & {\footnotesize{}GCE \cite{2018_NeurIPS_GCE}} & {\footnotesize{}DB \cite{2019_ICML_BynamicBootstrapping}} & {\footnotesize{}DMI \cite{2019_NeurIPS_LDMI}} & {\footnotesize{}P~\cite{2019_CVPR_JointOptimizImproved}} & {\footnotesize{}DM~\cite{2020_ICLR_DivideMix}} & {\footnotesize{}DRPL}\tabularnewline
\midrule 
{\footnotesize{}73.88} & {\footnotesize{}74.68} & {\footnotesize{}76.52} & {\footnotesize{}80.76} & {\footnotesize{}74.28} & {\footnotesize{}79.68} & {\footnotesize{}73.96} & {\footnotesize{}79.96} & {\footnotesize{}78.16} & \textbf{\footnotesize{}82.08}\tabularnewline
\bottomrule
\end{tabular}}
\end{table*}
\begin{table}[t]
\begin{centering}
\caption{\label{tab:CIFAReval}CIFAR-10(100) accuracy on top (bottom). Key:
NU (non-uniform noise). U (uniform noise). CE (Cross-entropy).}
\setlength{\tabcolsep}{3.0pt}
\global\long\def\arraystretch{0.9}%
{\footnotesize{}}%
\begin{tabular}{lcccccccc}
\toprule 
 & \multirow{2}{*}{{\footnotesize{}0\%}} & \multicolumn{3}{c}{{\footnotesize{}NU}} & \multicolumn{4}{c}{{\footnotesize{}U}}\tabularnewline
 &  & {\footnotesize{}10\%} & {\footnotesize{}30\%} & {\footnotesize{}40\%} & {\footnotesize{}20\%} & {\footnotesize{}40\%} & {\footnotesize{}60\%} & \multicolumn{1}{c}{{\footnotesize{}80\%}}\tabularnewline
\midrule 
\multirow{1}{*}{{\footnotesize{}CE}} & {\footnotesize{}93.87} & {\footnotesize{}90.97} & {\footnotesize{}90.22} & {\footnotesize{}88.16} & {\footnotesize{}87.75} & {\footnotesize{}83.32} & {\footnotesize{}75.71} & {\footnotesize{}43.69}\tabularnewline
\multirow{1}{*}{{\footnotesize{}FW \cite{2017_CVPR_ForwardLoss}}} & {\footnotesize{}94.61} & {\footnotesize{}90.85} & {\footnotesize{}87.95} & {\footnotesize{}84.85} & {\footnotesize{}85.46} & {\footnotesize{}80.67} & {\footnotesize{}70.86} & {\footnotesize{}45.58}\tabularnewline
\multirow{1}{*}{{\footnotesize{}R \cite{2018_CVPR_JointOpt}}} & {\footnotesize{}94.37} & {\footnotesize{}93.70} & {\footnotesize{}92.69} & {\footnotesize{}92.70} & {\footnotesize{}92.77} & {\footnotesize{}89.97} & {\footnotesize{}85.62} & {\footnotesize{}53.26}\tabularnewline
\multirow{1}{*}{{\footnotesize{}M \cite{2018_ICLR_mixup}}} & \textbf{\footnotesize{}96.07} & {\footnotesize{}93.79} & {\footnotesize{}91.38} & {\footnotesize{}87.01} & {\footnotesize{}91.27} & {\footnotesize{}85.84} & {\footnotesize{}80.78} & {\footnotesize{}57.93}\tabularnewline
\multirow{1}{*}{{\footnotesize{}GCE \cite{2018_NeurIPS_GCE}}} & {\footnotesize{}93.93} & {\footnotesize{}91.40} & {\footnotesize{}90.45} & {\footnotesize{}88.39} & {\footnotesize{}88.49} & {\footnotesize{}84.09} & {\footnotesize{}76.55} & {\footnotesize{}43.39}\tabularnewline
\multirow{1}{*}{{\footnotesize{}DB \cite{2019_ICML_BynamicBootstrapping}}} & {\footnotesize{}92.78} & {\footnotesize{}91.77} & {\footnotesize{}93.23} & {\footnotesize{}91.25} & {\footnotesize{}93.95} & {\footnotesize{}92.38} & \textbf{\footnotesize{}89.53} & {\footnotesize{}49.90}\tabularnewline
\multirow{1}{*}{{\footnotesize{}DMI \cite{2019_NeurIPS_LDMI}}} & {\footnotesize{}93.93} & {\footnotesize{}91.31} & {\footnotesize{}91.34} & {\footnotesize{}88.64} & {\footnotesize{}88.40} & {\footnotesize{}83.98} & {\footnotesize{}75.91} & {\footnotesize{}44.17}\tabularnewline
\multirow{1}{*}{{\footnotesize{}PEN \cite{2019_CVPR_JointOptimizImproved}}} & {\footnotesize{}93.94} & {\footnotesize{}93.19} & {\footnotesize{}92.94} & {\footnotesize{}91.63} & {\footnotesize{}92.87} & {\footnotesize{}91.34} & {\footnotesize{}89.15} & {\footnotesize{}56.14}\tabularnewline
\midrule 
\multirow{1}{*}{{\footnotesize{}DRPL}} & {\footnotesize{}94.47} & \textbf{\footnotesize{}95.70} & \textbf{\footnotesize{}93.65} & \textbf{\footnotesize{}93.14} & \textbf{\footnotesize{}94.20} & \textbf{\footnotesize{}92.92} & {\footnotesize{}89.21} & \textbf{\footnotesize{}64.35}\tabularnewline
\bottomrule
\end{tabular}{\footnotesize\par}
\par\end{centering}
\begin{centering}
\medskip{}
\par\end{centering}
\centering{}\setlength{\tabcolsep}{3.0pt}
\global\long\def\arraystretch{0.9}%
{\footnotesize{}}%
\begin{tabular}{lcccccccc}
\toprule 
 & \multirow{2}{*}{{\footnotesize{}0\%}} & \multicolumn{3}{c}{{\footnotesize{}NU}} & \multicolumn{4}{c}{{\footnotesize{}U}}\tabularnewline
 &  & {\footnotesize{}10\%} & {\footnotesize{}30\%} & {\footnotesize{}40\%} & {\footnotesize{}20\%} & {\footnotesize{}40\%} & {\footnotesize{}60\%} & \multicolumn{1}{c}{{\footnotesize{}80\%}}\tabularnewline
\midrule 
\multirow{1}{*}{{\footnotesize{}CE}} & {\footnotesize{}74.59} & {\footnotesize{}68.18} & {\footnotesize{}54.20} & {\footnotesize{}46.55} & {\footnotesize{}59.19} & {\footnotesize{}51.44} & {\footnotesize{}39.05} & {\footnotesize{}19.59}\tabularnewline
\multirow{1}{*}{{\footnotesize{}FW \cite{2017_CVPR_ForwardLoss}}} & {\footnotesize{}75.43} & {\footnotesize{}68.82} & {\footnotesize{}54.65} & {\footnotesize{}45.38} & {\footnotesize{}61.03} & {\footnotesize{}51.44} & {\footnotesize{}39.05} & {\footnotesize{}19.59}\tabularnewline
\multirow{1}{*}{{\footnotesize{}R \cite{2018_CVPR_JointOpt}}} & {\footnotesize{}74.43} & {\footnotesize{}73.09} & {\footnotesize{}68.25} & {\footnotesize{}59.49} & {\footnotesize{}70.79} & {\footnotesize{}66.35} & {\footnotesize{}57.48} & {\footnotesize{}30.65}\tabularnewline
\multirow{1}{*}{{\footnotesize{}M \cite{2018_ICLR_mixup}}} & \textbf{\footnotesize{}78.33} & {\footnotesize{}73.39} & {\footnotesize{}59.15} & {\footnotesize{}49.40} & {\footnotesize{}66.60} & {\footnotesize{}54.69} & {\footnotesize{}45.80} & {\footnotesize{}27.02}\tabularnewline
\multirow{1}{*}{{\footnotesize{}GCE \cite{2018_NeurIPS_GCE}}} & {\footnotesize{}74.91} & {\footnotesize{}68.34} & {\footnotesize{}55.56} & {\footnotesize{}47.24} & {\footnotesize{}60.09} & {\footnotesize{}61.23} & {\footnotesize{}49.75} & {\footnotesize{}25.77}\tabularnewline
\multirow{1}{*}{{\footnotesize{}DB \cite{2019_ICML_BynamicBootstrapping}}} & {\footnotesize{}70.64} & {\footnotesize{}68.19} & {\footnotesize{}62.81} & {\footnotesize{}55.76} & {\footnotesize{}69.12} & {\footnotesize{}64.84} & {\footnotesize{}57.85} & {\footnotesize{}46.45}\tabularnewline
\multirow{1}{*}{{\footnotesize{}DMI \cite{2019_NeurIPS_LDMI}}} & {\footnotesize{}74.75} & {\footnotesize{}68.29} & {\footnotesize{}54.40} & {\footnotesize{}46.65} & {\footnotesize{}59.16} & {\footnotesize{}53.49} & {\footnotesize{}41.49} & {\footnotesize{}20.50}\tabularnewline
\multirow{1}{*}{{\footnotesize{}PEN \cite{2019_CVPR_JointOptimizImproved}}} & {\footnotesize{}77.80} & \textbf{\footnotesize{}76.31} & {\footnotesize{}63.67} & {\footnotesize{}50.64} & \textbf{\footnotesize{}75.16} & {\footnotesize{}69.56} & {\footnotesize{}56.16} & {\footnotesize{}27.12}\tabularnewline
\midrule 
\multirow{1}{*}{{\footnotesize{}DRPL}} & {\footnotesize{}72.27} & {\footnotesize{}72.40} & \textbf{\footnotesize{}69.30} & \textbf{\footnotesize{}65.86} & {\footnotesize{}71.25} & \textbf{\footnotesize{}73.13} & \textbf{\footnotesize{}68.71} & \textbf{\footnotesize{}53.04}\tabularnewline
\bottomrule
\end{tabular}{\footnotesize\par}
\end{table}

\subsection{Other datasets and real-world frameworks\label{subsec:CIFAR_WebEval}}

\begin{figure}[t]
\centering{}\includegraphics[width=0.99\columnwidth]{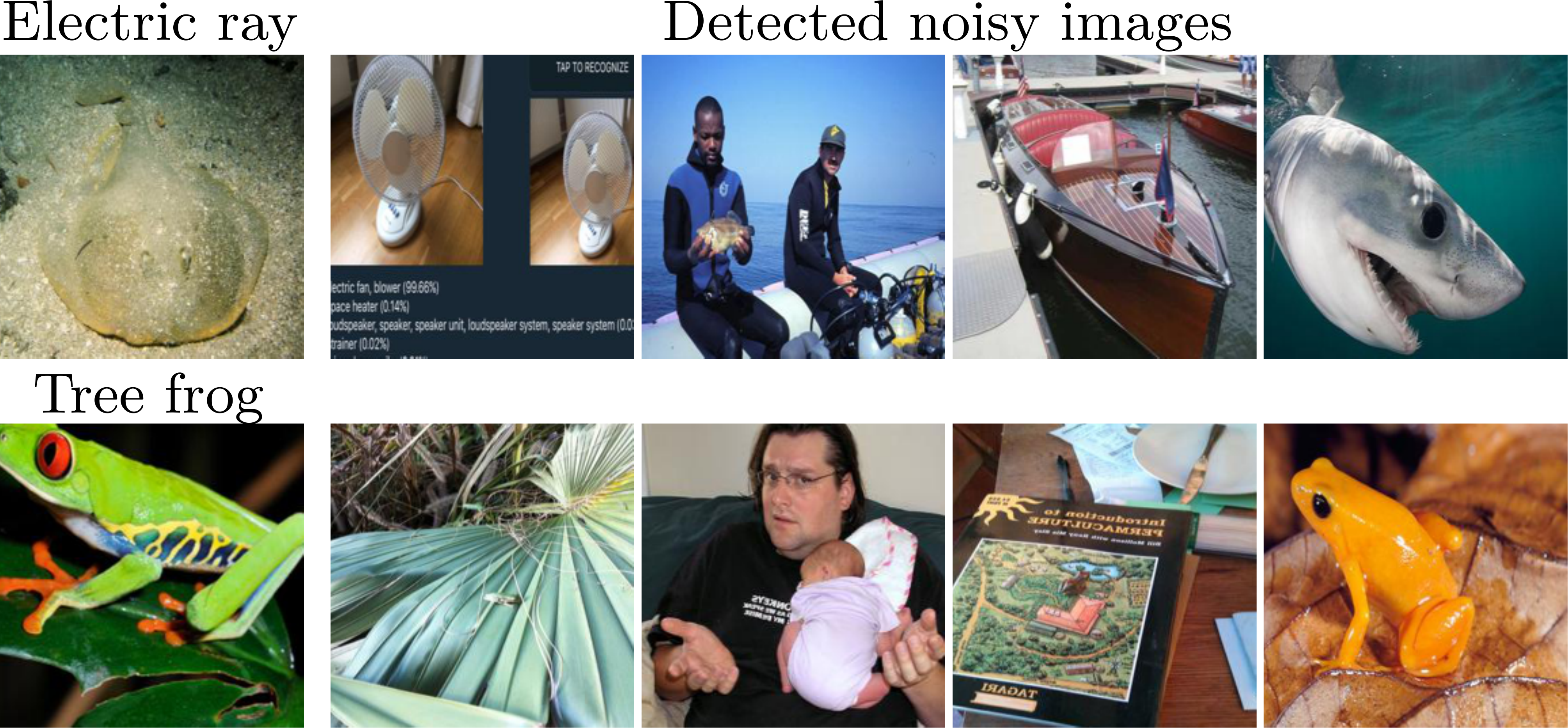}\caption{\label{fig:WebVisionImages} Examples of detected noisy images in
WebVision \cite{2017_arXiv_WebVision}. The first column shows a clean
example from the class.}
\end{figure}

Table \ref{tab:WebV} compares DRPL against related work in the first
50 classes of WebVision 1.0 dataset \cite{2017_arXiv_WebVision} to
verify the practical use of our method in real-world noise scenarios.
The proposed approach is more accurate than all compared approaches.
Note, however, that many approaches give very similar performance.
Surprisingly, a straightforward approach like M \cite{2018_ICLR_mixup},
not specifically designed to deal with label noise, gives the second
highest accuracy. The similar performance among approaches is something
that can be also observed for non-uniform OOD noise in ImageNet32/64.
Another similarity here with non-uniform OOD noise is the small degradation
in performance at the last training epoch (see supplementary material).
Furthermore, the reduced noise level in WebVision, estimated at around
20\% in \cite{2017_arXiv_WebVision}, and with 17\% of detected noisy
samples in our experiment, helps to explain performance similarities
across approaches: this is also seen in CIFAR and ImageNet32/64 with
low noise levels. We present detected noisy images in Figure \ref{fig:WebVisionImages}
(more examples in the supplementary material), which are mainly from
outside the distribution.

We also evaluate DRPL in CIFAR-10/100 (see Table~\ref{tab:CIFAReval}),
showing improved performance over all other compared approaches (extended
results in the supplementary material). We evaluate the same approaches
as in ImageNet32/64 \cite{2019_ICML_BynamicBootstrapping,2017_CVPR_ForwardLoss,2018_CVPR_JointOpt,2018_ICLR_mixup}
and also add further recent approaches \cite{2019_NeurIPS_LDMI,2019_CVPR_JointOptimizImproved,2018_NeurIPS_GCE}
(configuration details in supplementary material). Improved results
for uniform and non-uniform noise in CIFAR confirm the improvements
observed in ImageNet32/64. The additional recent approaches, GCE \cite{2018_NeurIPS_GCE},
DMI \cite{2019_NeurIPS_LDMI}, and PEN \cite{2019_CVPR_JointOptimizImproved}
are outperformed by our approach DRPL in most cases with GCE and DMI
far from top-performance. The exception is CIFAR-100 with little or
no noise where top performances of 78.33 for M \cite{2018_ICLR_mixup}
and 76.31 for PEN \cite{2019_CVPR_JointOptimizImproved} surpasses
DRPL, thus suggesting that the information discarded by our approach
is important for achieving top performance. This highlights that high
loss is indicative not only of incorrect labels, but also of difficult
samples \cite{2019_arXiv_BigLosers} and it is the semi-supervised approach
that determines whether this information is recovered.

Note that it is unfair to directly compare against numbers reported
in DM \cite{2020_ICLR_DivideMix} as they use a clean set to select
different hyperparameters for each noise-level and type. We prefer
not to use a clean set and adopt a common configuration, which makes
comparisons fairer. However, Table \ref{tab:WebV} reports
our run of DM (official code and configuration) in WebVision, where
there are no different noise types and levels, showing that the proposed DRPL
approach outperforms DM by 4 accuracy points. DM \cite{2020_ICLR_DivideMix}
also reports 77.32 accuracy results in a different WebVision setup
(subset of the first 50 classes and InceptionResNetv2 network). Our
approach under that setup achieves 77.12,  demonstrating that
the proposed approach is, at very least, competitive with the latest state-of-the-art.

It is worth noting that methods behave differently against artificially
introduced noise in CIFAR than in real-world scenarios like WebVision
\cite{2017_arXiv_WebVision}. Substantial accuracy degradation is
seen in many methods in CIFAR at the end of the training (see supplementary
material), while in WebVision degradations are minor, and mixup (M)
\cite{2018_ICLR_mixup}, which does not deal specifically with label
noise, outperforms most approaches. We believe that evaluation with
OOD noise in ImageNet32/64 helps in understanding how label noise
memorization might behave in real-world scenarios.

\section{Conclusion}

We propose a framework to study multiple label noise distributions
and a straightforward approach based on label noise detection and
semi-supervised learning to tackle them all.  We provide intuitions
about the network behavior when memorizing noisy labels and show that
such memorization does not often harm learning discriminative intermediate
representations. Results in five datasets support the generality and
robustness of our approach and help in reducing the gap between synthetic
and real-world label noise. Future work will study more noise distributions
and their combinations to further approach real-world noise.

\section*{Acknowledgement}

This publication has emanated from research conducted with the financial
support of Science Foundation Ireland (SFI) under grant number SFI/15/SIRG/3283
and SFI/12/RC/2289\_P2.

\bibliographystyle{IEEEtran}
\bibliography{refs}

\newpage{}

\section{Supplementary Material}

\subsection{Proposed method: algorithm}

Algorithm \ref{alg:AlgorithmSummary} summarizes the proposed label noise Distribution Robust Pseudo-Labeling (DRPL) approach to deal with label noise.

\begin{algorithm}
\SetKwInOut{Input}{input}\SetKwInOut{Output}{output}
\SetAlgoLined
\Input{Dataset $\mathcal{D}$ with potentially noisy labels $y$.}
\Output{Model $h_{\theta}(x)$, clean $\mathcal{D}_{l}$ and noisy $\mathcal{D}_{u}$ samples.}
\tcp{Noise detection: first stage}
 Relabeling: train model $h_{\phi}(x)$ on $\mathcal{D}$ using Eq. 1\;
 Cross-entropy loss $\ell^{*}\left(\phi\right)$ w.r.t. original labels $y$.\;
 Initial noise model: fit two-component BMM to $\ell^{*}\left(\phi\right)$.\;
 Initial labeled/unlabeled sets: create $\mathcal{D}_{l}^{\prime}$ and $\mathcal{D}_{u}^{\prime}$ using the initial BMM (Eqs. 2 and 3).\;
 \tcp{Noise detection: second stage}
 Semi-supervised training of $h_{\varphi}(x)$ with $\mathcal{D}_{l}^{\prime}$ and $\mathcal{D}_{u}^{\prime}$.\;
 Cross-entropy loss $\ell^{*}\left(\varphi\right)$ w.r.t. original labels $y$.\;
 Final noise model: fit two-component BMM to $\ell^{*}\left(\varphi\right)$.\;
 Final labeled/unlabeled sets: create $\mathcal{D}_{l}$ and $\mathcal{D}_{u}$ using the final BMM as in Eqs. 2 and 3.\;
 \tcp{Semi-supervised learning}
 Semi-supervised training of $h_{\theta}(x)$ with $\mathcal{D}_{l}$ and $\mathcal{D}_{u}$.\;
 
 \caption{\label{alg:AlgorithmSummary} Summary of proposed method}
\end{algorithm}

\subsection{SSL vs label noise transition matrix}

Correcting the loss by using the label noise transition matrix as
proposed in \cite{2017_CVPR_ForwardLoss} has recently attracted a
lot of interest \cite{2018_NIPS_GoldLoss,2019_AAAI_Safeguard}.
Estimating the label noise transition matrix $T$, however, is a challenging
task as label flips from one class to another have to be estimated.
It seems simpler, on the other hand, to detect clean and noisy samples and
discard the noisy labels. 

Table \ref{tab:PerformanceOracles} presents a study using oracles
for both tasks (i.e. perfect knowledge clean-noisy samples and known $T$) to shed
light the potential of each approach under ideal conditions. The results
show that SSL surpasses label noise transition matrix correction \cite{2017_CVPR_ForwardLoss}
for both uniform and non-uniform noise. We believe this is an important
observation that suggests further research on making the label noise transition matrix methods more effective.

\subsection{Extended implementation details\label{subsec:ExtendedDetails-1}}

The $\lambda_{1}$ and $\lambda_{2}$ hyperparameters
are set as in \cite{2018_CVPR_JointOpt}; we have not sought careful
tuning. We use the default parameters for the semi-supervised method
\cite{2019_arXiv_Pseudo}. Two important hyperparameters in our method
are the thresholds $\gamma_{1}$ and $\gamma_{2}$ to detect label
noise in each label noise detection stage. In the first stage, the
idea is to get sufficient data for semi-supervised learning with labels that are as
clean as possible. We select $\gamma_{1}=0.05$ in ImageNet32/64
and WebVision, and $\gamma_{1}=0.1$ in CIFAR. We  slightly
increase the threshold in CIFAR to assure that enough data is selected
to perform a successful semi-supervised learning, as we had some problems in CIFAR-100 to select enough data
(less than 4K samples were selected occasionally in CIFAR-100, which
according to results in \cite{2019_arXiv_Pseudo} is not enough to
prevent performance degradation). We keep the
same configuration in CIFAR-10 to demonstrate its generality.

\begin{table}[t]
\begin{centering}
\caption{\label{tab:PerformanceOracles}ImageNet32/64 accuracy using oracles.}
\medskip{}
\par\end{centering}
\centering{}\renewcommand{\arraystretch}{1}{\small{}}%
\begin{tabular}{lcc}
\toprule
 & {\small{}Oracle SSL} & {\small{}Oracle forward}\tabularnewline
\midrule 
\multirow{1}{*}{{\small{}Uniform ID (80\%)}} & \textbf{\small{}69.06 / 78.02} & {\small{}34.66 / 49.98}\tabularnewline
\multirow{1}{*}{{\small{}Non-uniform ID (50\%)}} & \textbf{\small{}73.24 / 81.80} & {\small{}65.70 / 73.44}\tabularnewline
\bottomrule 
\end{tabular}{\small{}}{\small\par}
\end{table}

Training details for compared methods: 
\begin{itemize}
\item F \cite{2017_CVPR_ForwardLoss}: In ImageNet32/64 and CIFAR we train
200 epochs with initial learning rate of 0.1 that we divide by 10
in epochs 100 and 150. For WebVision we train 125 epochs and reduce
the learning rate in epochs 75 and 120. Forward correction always
starts in epoch 50. 
\item M \cite{2018_ICLR_mixup}: In ImageNet32/64 and CIFAR we train 300
epochs with initial learning rate of 0.1 that we divide by 10 in epochs
100 and 250. For WebVision we train 200 epochs and reduce the learning
rate in epochs 75 and 120. Mixup parameter $\alpha$ is set to 1. 
\item R \cite{2018_CVPR_JointOpt}: In ImageNet32/64 and CIFAR we train
in the first stage of the method as in the first stage of our label
noise detection method, i.e. we train for 100 epochs with initial
learning rate of 0.1 that we divide by 10 in epochs 45 and 80. Relabeling
starts in epoch 40. For the second stage we train 120 epochs with
initial learning rate of 0.1 that we divide by 10 in epochs 40 and
80. Other hyperparameters are kept as in  \cite{2018_CVPR_JointOpt}. 
\item DB \cite{2019_ICML_BynamicBootstrapping}: We use the code\footnote{\url{https://git.io/fjsvE}}
associated to the official implementation of \cite{2019_ICML_BynamicBootstrapping}.
We keep the default configuration used for CIFAR-10/100 and use it
also in ImageNet32/64. For WebVision we train for 200 epochs with
initial learning rate of 0.1 that we divide by 10 in epochs 100 and
175 (bootstrapping starts in epoch 102). 
\item GCE and DMI \cite{2019_NeurIPS_LDMI}: We use the code\footnote{\url{https://git.io/JeRGh}}
associated to the official implementation of \cite{2019_NeurIPS_LDMI}.
We keep the default configuration used for CIFAR-10 for both CIFAR-10
and CIFAR-100. We respect the use of the best model in the pre-training
phase using cross-entropy loss, although such selection would not
be straightforward without using a clean set in a real scenario.

\item P \cite{2019_CVPR_JointOptimizImproved}: We use the official implementation\footnote{\url{https://git.io/JezIO}}
of \cite{2019_CVPR_JointOptimizImproved}. We found difficulties on configuring this method for the different datasets, as similar configurations did not work in CIFAR-10/100, and WebVision. We always use $\alpha=0.1$, $\beta=0.4$, and learning rate of 0.1
in the first and second stages, whereas we use 0.2 as starting learning
rate in the third stage. For CIFAR-10 different hyperparameters are
used in \cite{2019_CVPR_JointOptimizImproved} to deal with different
noise distributions and noise levels, which we did not find reasonable.
Therefore, we set a common configuration with
$\lambda=600$ and training the suggested number of epochs \cite{2019_CVPR_JointOptimizImproved}.
In CIFAR-100 we tried the configuration suggested in the paper and
it did not converge to reasonable performance, thus we reduced the
learning rate from 0.35 to 0.1 to make it stable and use $\lambda=10000$
as suggested. For WebVision we used the same configuration used in
CIFAR-10 and trained 40 epochs in the first stage, 60 in the second
and 100 in the third. Note that we tried the suggested CIFAR-100 configuration in WebVision, but it led to poor performance.
\end{itemize}



\begin{figure}[t]
\centering{}\setlength{\tabcolsep}{0.1pt}\resizebox{1.0\columnwidth}{!}{%
\begin{tabular}{cc}
\includegraphics[width=0.24\textwidth]{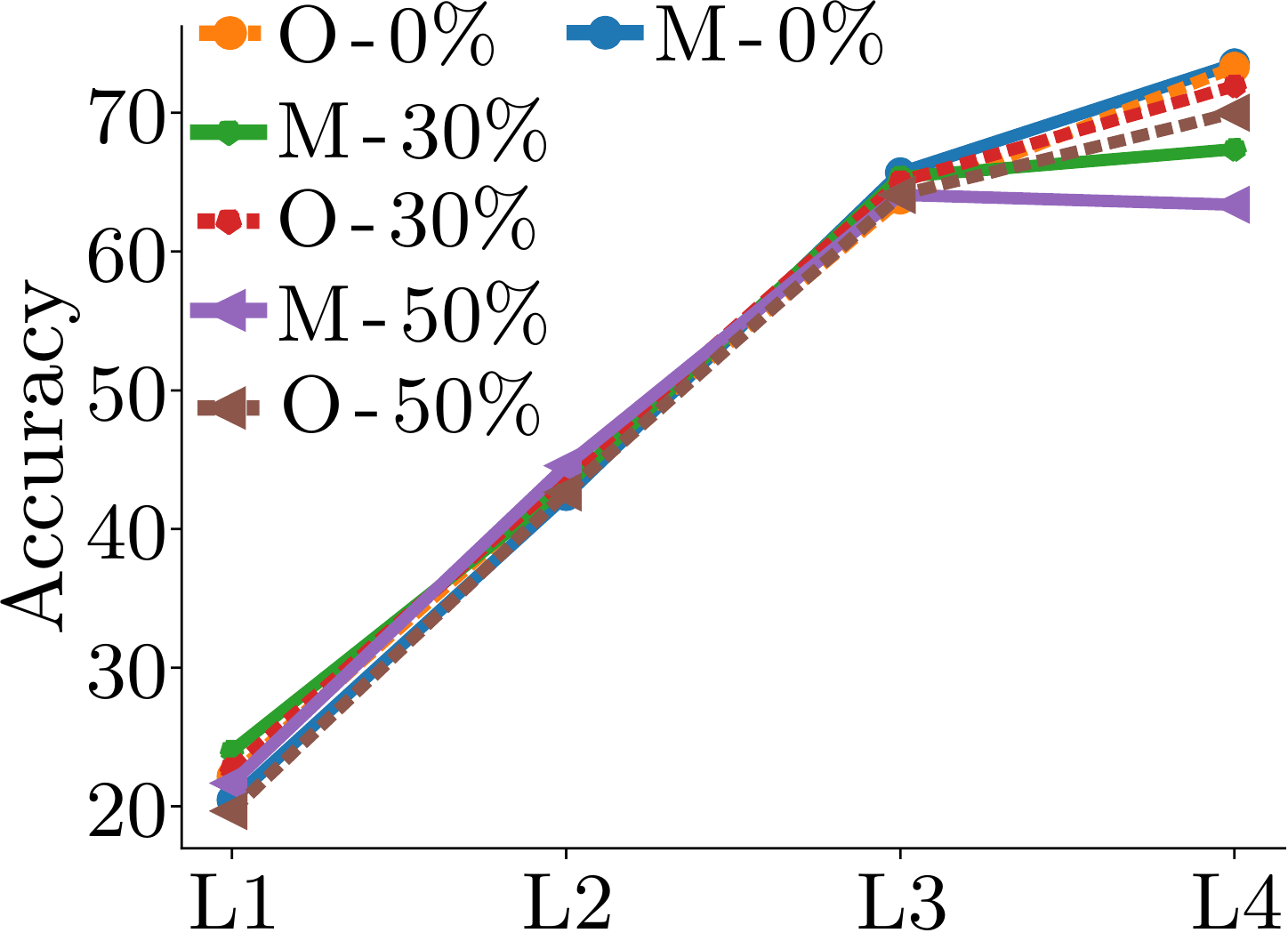}  & \includegraphics[width=0.24\textwidth]{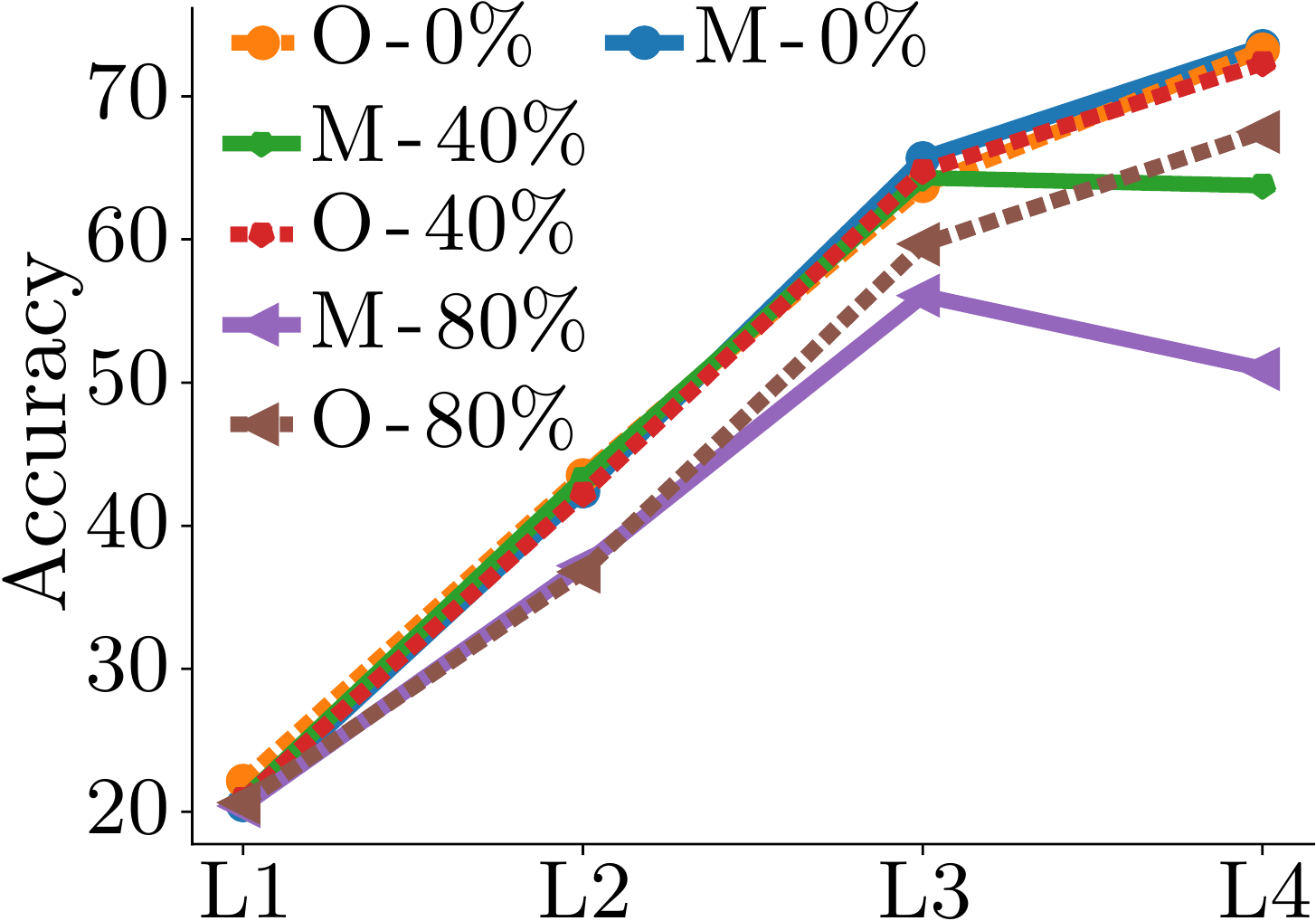} \tabularnewline
(a)  & (b) \tabularnewline
\includegraphics[width=0.24\textwidth]{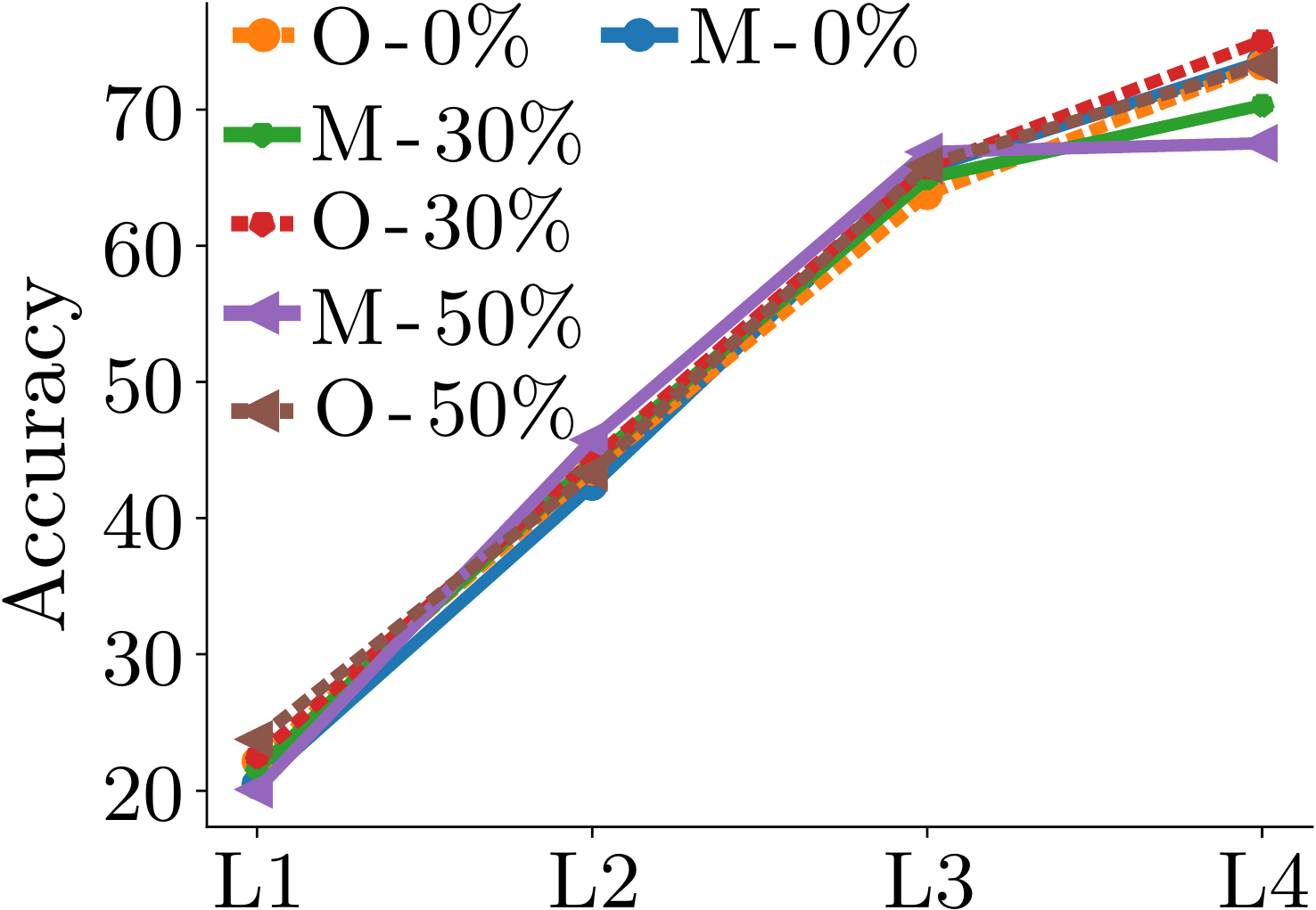}  & \includegraphics[width=0.24\textwidth]{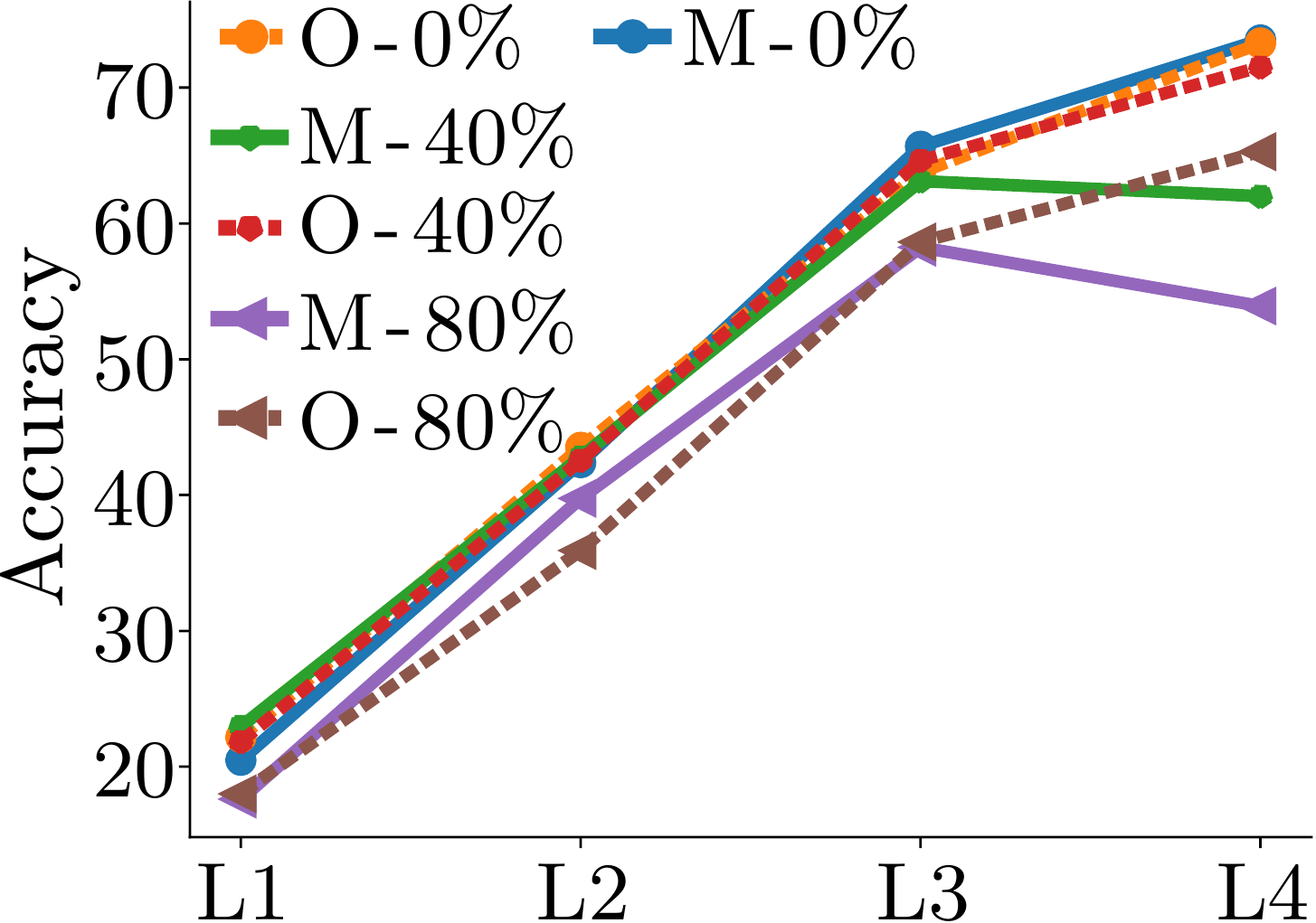}\tabularnewline
(c)  & (d)\tabularnewline
\end{tabular}}\caption{\label{fig:TransferFigures-1-1} ImageNet32 linear probes. A linear
classifier is trained using features at different depths. Source domain:
100 classes. Target domain: 25 classes from the remaining 900. Noise:
in-distribution non-uniform (a) and uniform (b), and out-of-distribution
non-uniform (c) and uniform (d). Model from the last training epoch
in the source domain is used (i.e. M has fitted the noise). Key: M:
Mixup. O: Ours.}
\end{figure}

Figure \ref{fig:TransferFigures-1-1} reports the results of the linear probes experiment in ImageNet32, showing similar characteristics as those in ImageNet64 reported in the paper. Similar conclusions can be extracted, highlighting little or no degradation of intermediate features when learning with label noise.

\subsection{Extended results: ImageNet32/64\label{subsec:ExtendedResultsIm-1}}

Tables \ref{tab:AppendixIm32-1} and \ref{tab:AppendixIm64-1}
 report the extended results for ImageNet32/64 with four different label noise distributions, more noise levels, and include accuracy in the last training epoch (useful to understand when methods fit label noise).

\subsection{Extended results: CIFAR-10/100\label{subsec:ExtendedResultsCIF-1}}

Table \ref{tab:ExtendedCIFAReval-1} gives additional results for
CIFAR-10/100 adding the performance in the last epoch, which as found
in ImageNet32/64, reveals methods degradation due to label noise memorization
(e.g. CE, M).

\begin{table*}
\centering{}\caption{\label{tab:AppendixIm32-1}ImageNet32 accuracy for 100 classes. Bold
denotes best performance. Key: NU (Non-uniform). U (Uniform). ID (in-distribution).
OOD (out-of-distribution).}
\medskip{}
\renewcommand{\arraystretch}{1}\resizebox{1\textwidth}{!}{{\small{}}%
\begin{tabular}{ccccccccccccccccc}
\toprule
 &  & \multirow{2}{*}{0{\small{}\%}} & \multicolumn{3}{c|}{NU-ID} & \multicolumn{4}{c|}{U-ID} & \multicolumn{3}{c|}{NU-OOD} & \multicolumn{4}{c}{U-OOD}\tabularnewline
 &  &  & 10{\small{}\%} & 30{\small{}\%} & 50{\small{}\%} & 20{\small{}\%} & 40{\small{}\%} & 60{\small{}\%} & 80{\small{}\%} & 10{\small{}\%} & 30{\small{}\%} & 50{\small{}\%} & 20{\small{}\%} & 40{\small{}\%} & 60{\small{}\%} & 80{\small{}\%}\tabularnewline
\midrule
\multirow{2}{*}{FW \cite{2017_CVPR_ForwardLoss}} & \emph{Best} & {\small{}71.84} & {\small{}65.84} & {\small{}54.22} & {\small{}43.38} & {\small{}57.40} & {\small{}52.06} & {\small{}44.36} & {\small{}31.20} & {\small{}67.50} & {\small{}62.14} & {\small{}55.06} & {\small{}63.04} & {\small{}56.32} & {\small{}50.16} & {\small{}40.08}\tabularnewline
 & \emph{Last} & {\small{}71.42} & {\small{}65.84} & {\small{}52.00} & {\small{}33.88} & {\small{}57.30} & {\small{}42.48} & {\small{}25.28} & {\small{}11.02} & {\small{}67.20} & {\small{}61.90} & {\small{}54.00} & {\small{}62.80} & {\small{}53.58} & {\small{}42.86} & {\small{}26.42}\tabularnewline
\multirow{2}{*}{R \cite{2018_CVPR_JointOpt}} & \emph{Best} & {\small{}69.26} & {\small{}68.64} & {\small{}67.24} & {\small{}63.62} & {\small{}66.66} & {\small{}62.98} & {\small{}57.54} & {\small{}41.52} & {\small{}68.28} & {\small{}66.36} & {\small{}62.80} & {\small{}67.20} & {\small{}64.04} & {\small{}58.92} & {\small{}45.00}\tabularnewline
 & \emph{Last} & {\small{}69.06} & {\small{}68.40} & {\small{}66.98} & {\small{}63.22} & {\small{}66.30} & {\small{}62.56} & {\small{}57.34} & {\small{}41.24} & {\small{}67.96} & {\small{}66.28} & {\small{}62.48} & {\small{}67.20} & {\small{}63.86} & {\small{}58.54} & {\small{}44.64}\tabularnewline

\multirow{2}{*}{M \cite{2018_ICLR_mixup}} & \emph{Best} & \textbf{\small{}74.80} & {\small{}71.64} & {\small{}67.14} & {\small{}51.96} & {\small{}67.76} & {\small{}61.98} & {\small{}53.84} & {\small{}38.92} & {\small{}71.10} & {\small{}66.14} & {\small{}60.62} & {\small{}69.88} & {\small{}64.66} & {\small{}58.48} & {\small{}47.40}\tabularnewline
 & \emph{Last} & \textbf{\small{}74.80} & {\small{}69.74} & {\small{}57.00} & {\small{}37.56} & {\small{}62.72} & {\small{}48.40} & {\small{}31.30} & {\small{}12.32} & {\small{}70.76} & {\small{}64.68} & {\small{}56.38} & {\small{}65.50} & {\small{}56.80} & {\small{}44.84} & {\small{}26.86}\tabularnewline

\multirow{2}{*}{DB \cite{2019_ICML_BynamicBootstrapping}} & \emph{Best} & {\small{}66.18} & {\small{}67.54} & {\small{}62.88} & {\small{}52.20} & {\small{}68.42} & {\small{}67.62} & {\small{}63.56} & {\small{}45.34} & {\small{}67.40} & {\small{}64.86} & {\small{}60.58} & {\small{}67.28} & {\small{}65.96} & {\small{}59.26} & {\small{}39.30}\tabularnewline
 & \emph{Last} & {\small{}65.42} & {\small{}66.48} & {\small{}61.88} & {\small{}51.64} & {\small{}67.84} & {\small{}62.98} & {\small{}62.72} & {\small{}44.56} & {\small{}66.90} & {\small{}64.08} & {\small{}60.18} & {\small{}66.72} & {\small{}65.28} & {\small{}58.86} & {\small{}38.44}\tabularnewline
\midrule
\multirow{2}{*}{DRPL (ours)} & \emph{Best} & 72.98 & \textbf{72.60} & \textbf{73.46} & \textbf{68.18} & \textbf{74.28} & \textbf{73.48} & \textbf{70.06} & \textbf{61.78} & \textbf{72.60} & \textbf{71.38} & \textbf{67.32} & \textbf{73.72} & \textbf{71.36} & \textbf{65.52} & \textbf{54.10}\tabularnewline
 & \emph{Last} & 72.60 & \textbf{72.34} & \textbf{73.00} & \textbf{67.56} & \textbf{73.80} & \textbf{73.30} & \textbf{70.06} & \textbf{61.30} & \textbf{72.26} & \textbf{71.36} & \textbf{67.32} & \textbf{73.20} & \textbf{70.64} & \textbf{65.18} & \textbf{53.16}\tabularnewline
\bottomrule
\end{tabular}}
\end{table*}

\begin{table*}
\centering{}\caption{\label{tab:AppendixIm64-1}ImageNet64 accuracy for 100 classes. Bold
denotes best performance. Key: NU (Non-uniform). U (Uniform). ID (in-distribution).
OOD (out-of-distribution).}
\medskip{}
\renewcommand{\arraystretch}{1}\resizebox{1\textwidth}{!}{{\small{}}%
\begin{tabular}{ccccccccccccccccc}
\toprule
 &  & \multirow{2}{*}{{\small{}0\%}} & \multicolumn{3}{c|}{{\small{}NU-ID}} & \multicolumn{4}{c|}{{\small{}U-ID}} & \multicolumn{3}{c|}{{\small{}NU-OOD}} & \multicolumn{4}{c}{{\small{}U-OOD}}\tabularnewline

 &  &  & {\small{}10\%} & {\small{}30\%} & {\small{}50\%} & {\small{}20\%} & {\small{}40\%} & {\small{}60\%} & {\small{}80\%} & {\small{}10\%} & {\small{}30\%} & {\small{}50\%} & {\small{}20\%} & {\small{}40\%} & {\small{}60\%} & {\small{}80\%}\tabularnewline
\midrule
\multirow{2}{*}{{\small{}FW \cite{2017_CVPR_ForwardLoss}}} & \emph{\small{}Best} & {\small{}79.80} & {\small{}73.08} & {\small{}60.10} & {\small{}46.06} & {\small{}64.06} & {\small{}57.42} & {\small{}51.52} & {\small{}37.84} & {\small{}73.08} & {\small{}69.86} & {\small{}63.38} & {\small{}70.86}  & {\small{}63.08} & {\small{}58.14}  & {\small{}47.68} \tabularnewline
 & \emph{\small{}Last} & {\small{}79.32} & {\small{}73.08} & {\small{}60.10} & {\small{}42.96} & {\small{}63.92} & {\small{}49.70} & {\small{}31.28} & {\small{}14.18} & {\small{}73.08} & {\small{}69.42} & {\small{}62.50} & {\small{}70.60}  & {\small{}61.78} & {\small{}51.74} & {\small{}35.56} \tabularnewline
\multirow{2}{*}{{\small{}R \cite{2018_CVPR_JointOpt}}} & \emph{\small{}Best} & {\small{}76.46} & {\small{}75.94} & {\small{}74.28} & {\small{}69.20} & {\small{}73.70} & {\small{}70.98} & {\small{}65.08} & {\small{}48.44} & {\small{}75.90} & {\small{}74.22}  & {\small{}70.74} & {\small{}75.54}  & {\small{}72.78}  & {\small{}66.90}  & {\small{}54.00}\tabularnewline
 & \emph{\small{}Last} & {\small{}76.26} & {\small{}75.28} & {\small{}74.02} & {\small{}68.84} & {\small{}73.40} & {\small{}70.74} & {\small{}65.04} & {\small{}48.28} & {\small{}75.48} & {\small{}73.70}  & {\small{}70.56} & {\small{}75.22}  & {\small{}72.48}  & {\small{}66.64}  & {\small{}53.72}\tabularnewline
\multirow{2}{*}{{\small{}M \cite{2018_ICLR_mixup}}} & \emph{\small{}Best} & \textbf{\small{}83.38} & {\small{}80.08} & {\small{}74.02} & {\small{}58.14} & {\small{}76.34} & {\small{}69.90} & {\small{}60.96} & {\small{}49.22} & {\small{}79.88} & {\small{}74.78} & {\small{}69.40} & {\small{}78.44} & {\small{}73.94} & {\small{}67.86} & {\small{}59.54}\tabularnewline
 & \emph{\small{}Last} & \textbf{\small{}82.88} & {\small{}78.42} & {\small{}63.54} & {\small{}44.56} & {\small{}70.86} & {\small{}54.60} & {\small{}34.12} & {\small{}14.12} & {\small{}79.14} & {\small{}72.92} & {\small{}64.76} & {\small{}73.26} & {\small{}64.00} & {\small{}51.16} & {\small{}33.28}\tabularnewline
\multirow{2}{*}{{\small{}DB \cite{2019_ICML_BynamicBootstrapping}}} & \emph{\small{}Best} & {\small{}76.16} & {\small{}75.38} & {\small{}71.30} & {\small{}60.98} & {\small{}75.86} & {\small{}74.56} & {\small{}71.50} & {\small{}56.44} & {\small{}80.92} & {\small{}77.94} & {\small{}70.38} & {\small{}75.24}  & {\small{}74.08} & {\small{}68.32} & {\small{}50.98}\tabularnewline
 & \emph{\small{}Last} & {\small{}71.48} & {\small{}72.70} & {\small{}70.66} & {\small{}58.70} & {\small{}74.38} & {\small{}73.76} & {\small{}70.00} & {\small{}55.74} & {\small{}80.76} & {\small{}77.96} & {\small{}70.24} & {\small{}72.54}  & {\small{}73.52} & {\small{}67.50}  & {\small{}50.78}\tabularnewline
\hline 
\multirow{2}{*}{{\small{}DRPL (ours)}} & \emph{\small{}Best} & {\small{}81.22} & \textbf{\small{}81.84} & \textbf{\small{}81.90} & \textbf{\small{}77.66} & \textbf{\small{}82.74} & \textbf{\small{}81.50} & \textbf{\small{}79.66} & \textbf{\small{}73.08} & \textbf{\small{}82.12} & \textbf{\small{}80.44} & \textbf{\small{}76.38} & \textbf{\small{}82.78} & \textbf{\small{}79.76} & \textbf{\small{}75.72} & \textbf{\small{}64.24}\tabularnewline
 & \emph{\small{}Last} & {\small{}80.78} & \textbf{\small{}81.54} & \textbf{\small{}81.40} & \textbf{\small{}77.08} & \textbf{\small{}82.20} & \textbf{\small{}81.20} & \textbf{\small{}79.28} & \textbf{\small{}72.60} & \textbf{\small{}81.80} & \textbf{\small{}80.34} & \textbf{\small{}75.90} & \textbf{\small{}82.40} & \textbf{\small{}79.46} & \textbf{\small{}75.18} & \textbf{\small{}63.28}\tabularnewline
\bottomrule 
\end{tabular}}
\end{table*}

\begin{table*}
\centering{}\caption{\label{tab:ExtendedCIFAReval-1}CIFAR-10/100 accuracy. Key: NU (non-uniform
noise). U (uniform noise). CE: Cross-entropy. Best (last) denotes
the accuracy obtained in the best (last) epoch.}
\medskip{}
\setlength{\tabcolsep}{3pt}\renewcommand{\arraystretch}{1}\resizebox{0.93\textwidth}{!}{{\small{}}%

\begin{tabular}{llcccccccccccccccc}
\toprule
 &  & \multicolumn{8}{c|}{{\small{}CIFAR-10}} & \multicolumn{8}{c}{{\small{}CIFAR-100}}\tabularnewline

 &  & \multirow{2}{*}{{\small{}0\%}} & \multicolumn{3}{c|}{{\small{}NU}} & \multicolumn{4}{c|}{{\small{}U}} & \multirow{2}{*}{{\small{}0\%}} & \multicolumn{3}{c|}{{\small{}NU}} & \multicolumn{4}{c}{{\small{}U}}\tabularnewline
 
 &  &  & {\small{}10\%} & {\small{}30\%} & {\small{}40\%} & {\small{}20\%} & {\small{}40\%} & {\small{}60\%} & \multicolumn{1}{c|}{{\small{}80\%}} &  & {\small{}10\%} & {\small{}30\%} & {\small{}40\%} & {\small{}20\%} & {\small{}40\%} & {\small{}60\%} & \multicolumn{1}{c}{\small{}80\%} \tabularnewline

\midrule 
\multirow{2}{*}{{\small{}CE}} & \emph{\small{}Best} & {\small{}93.87} & {\small{}90.97} & {\small{}90.22} & {\small{}88.16} & {\small{}87.75} & {\small{}83.32} & {\small{}75.71} & {\small{}43.69} & {\small{}74.59} & {\small{}68.18} & {\small{}54.20} & {\small{}46.55} & {\small{}59.19} & {\small{}51.44} & {\small{}39.05} & {\small{}19.59}\tabularnewline
 & \emph{\small{}Last} & {\small{}93.85} & {\small{}88.81} & {\small{}81.69} & {\small{}76.04} & {\small{}78.93} & {\small{}55.06} & {\small{}36.75} & {\small{}33.09} & {\small{}74.34} & {\small{}68.10} & {\small{}53.28} & {\small{}44.46} & {\small{}58.75} & {\small{}42.92} & {\small{}25.33} & {\small{}8.29}\tabularnewline
\multirow{2}{*}{{\small{}FW \cite{2017_CVPR_ForwardLoss}}} & \emph{\small{}Best} & {\small{}94.61} & {\small{}90.85} & {\small{}87.95} & {\small{}84.85} & {\small{}85.46} & {\small{}80.67} & {\small{}70.86} & {\small{}45.58} & {\small{}75.43} & {\small{}68.82} & {\small{}54.65} & {\small{}45.38} & {\small{}61.03} & {\small{}51.44} & {\small{}39.05} & {\small{}19.59}\tabularnewline
 & \emph{\small{}Last} & {\small{}94.45} & {\small{}90.39} & {\small{}81.87} & {\small{}76.65} & {\small{}81.03} & {\small{}61.06} & {\small{}42.16} & {\small{}24.09} & {\small{}75.37} & {\small{}68.66} & {\small{}54.42} & {\small{}45.32} & {\small{}60.83} & {\small{}45.31} & {\small{}26.79} & {\small{}8.44}\tabularnewline
\multirow{2}{*}{{\small{}R \cite{2018_CVPR_JointOpt}}} & \emph{\small{}Best} & {\small{}94.37} & {\small{}93.70} & {\small{}92.69} & {\small{}92.70} & {\small{}92.77} & {\small{}89.97} & {\small{}85.62} & {\small{}53.26} & {\small{}74.43} & {\small{}73.09} & {\small{}68.25} & {\small{}59.49} & {\small{}70.79} & {\small{}66.35} & {\small{}57.48} & {\small{}30.65}\tabularnewline
 & \emph{\small{}Last} & {\small{}94.21} & {\small{}93.61} & {\small{}92.52} & {\small{}92.54} & {\small{}92.58} & {\small{}89.96} & {\small{}85.42} & {\small{}52.96} & {\small{}74.20} & {\small{}72.71} & {\small{}68.13} & {\small{}59.41} & {\small{}70.70} & {\small{}66.18} & {\small{}57.22} & {\small{}30.53}\tabularnewline
\multirow{2}{*}{{\small{}M \cite{2018_ICLR_mixup}}} & \emph{\small{}Best} & \textbf{\small{}96.07} & {\small{}93.79} & {\small{}91.38} & {\small{}87.01} & {\small{}91.27} & {\small{}85.84} & {\small{}80.78} & {\small{}57.93} & \textbf{\small{}78.33} & {\small{}73.39} & {\small{}59.15} & {\small{}49.40} & {\small{}66.60} & {\small{}54.69} & {\small{}45.80} & {\small{}27.02}\tabularnewline
 & \emph{\small{}Last} & {\small{}95.96} & {\small{}93.30} & {\small{}83.26} & {\small{}77.74} & {\small{}84.76} & {\small{}66.07} & {\small{}43.95} & {\small{}20.38} & {\small{}77.90} & {\small{}72.40} & {\small{}57.63} & {\small{}48.07} & {\small{}66.40} & {\small{}52.20} & {\small{}33.73} & {\small{}13.21}\tabularnewline
\multirow{2}{*}{{\small{}GCE \cite{2018_NeurIPS_GCE}}} & \emph{\small{}Best} & {\small{}93.93} & {\small{}91.40} & {\small{}90.45} & {\small{}88.39} & {\small{}88.49} & {\small{}84.09} & {\small{}76.55} & {\small{}43.39} & {\small{}74.91} & {\small{}68.34} & {\small{}55.56} & {\small{}47.24} & {\small{}60.09} & {\small{}61.23} & {\small{}49.75} & {\small{}25.77}\tabularnewline
 & \emph{\small{}Last} & {\small{}93.85} & {\small{}89.48} & {\small{}80.60} & {\small{}76.04} & {\small{}82.95} & {\small{}80.84} & {\small{}66.95} & {\small{}20.19} & {\small{}74.68} & {\small{}68.01} & {\small{}52.84} & {\small{}45.02} & {\small{}59.49} & {\small{}55.50} & {\small{}39.96} & {\small{}21.65}\tabularnewline
\multirow{2}{*}{{\small{}DB \cite{2019_ICML_BynamicBootstrapping}}} & \emph{\small{}Best} & {\small{}92.78} & {\small{}91.77} & {\small{}93.23} & {\small{}91.25} & {\small{}93.95} & {\small{}92.38} & \textbf{\small{}89.53} & {\small{}49.90} & {\small{}70.64} & {\small{}68.19} & {\small{}62.81} & {\small{}55.76} & {\small{}69.12} & {\small{}64.84} & {\small{}57.85} & {\small{}46.45}\tabularnewline
 & \emph{\small{}Last} & {\small{}79.18} & {\small{}89.58} & {\small{}92.20} & {\small{}91.20} & {\small{}93.82} & {\small{}92.26} & \textbf{\small{}89.15} & {\small{}15.53} & {\small{}64.79} & {\small{}67.09} & {\small{}58.59} & {\small{}47.44} & {\small{}69.11} & {\small{}62.78} & {\small{}52.33} & {\small{}45.67}\tabularnewline
\multirow{2}{*}{{\small{}DMI \cite{2019_NeurIPS_LDMI}}} & \emph{\small{}Best} & {\small{}93.93} & {\small{}91.31} & {\small{}91.34} & {\small{}88.64} & {\small{}88.40} & {\small{}83.98} & {\small{}75.91} & {\small{}44.17} & {\small{}74.75} & {\small{}68.29} & {\small{}54.40} & {\small{}46.65} & {\small{}59.16} & {\small{}53.49} & {\small{}41.49} & {\small{}20.50}\tabularnewline
 & \emph{\small{}Last} & {\small{}93.88} & {\small{}91.11} & {\small{}91.16} & {\small{}83.99} & {\small{}88.33} & {\small{}83.24} & {\small{}14.78} & {\small{}43.67} & {\small{}74.44} & {\small{}68.15} & {\small{}54.15} & {\small{}46.20} & {\small{}58.82} & {\small{}53.22} & {\small{}41.29} & {\small{}20.30}\tabularnewline
\multirow{2}{*}{{\small{}PEN \cite{2019_CVPR_JointOptimizImproved}}} & \emph{\small{}Best} & {\small{}93.94} & {\small{}93.19} & {\small{}92.94} & {\small{}91.63} & {\small{}92.87} & {\small{}91.34} & {\small{}89.15} & {\small{}56.14} & {\small{}77.80} & \textbf{\small{}76.31} & {\small{}63.67} & {\small{}50.64} & \textbf{\small{}75.16} & {\small{}69.56} & {\small{}56.16} & {\small{}27.12}\tabularnewline
 & \emph{\small{}Last} & {\small{}93.89} & {\small{}93.14} & {\small{}92.85} & {\small{}91.57} & {\small{}92.72} & {\small{}91.32} & {\small{}89.05} & {\small{}55.99} & {\small{}77.75} & {\small{}76.05} & {\small{}59.29} & {\small{}48.26} & {\small{}74.93} & {\small{}68.49} & {\small{}53.08} & {\small{}25.41}\tabularnewline
\hline 
\multirow{2}{*}{{\small{}DRPL (ours)}} & \emph{\small{}Best} & {\small{}94.47} & \textbf{\small{}95.70} & \textbf{\small{}93.65} & \textbf{\small{}93.14} & \textbf{\small{}94.20} & \textbf{\small{}92.92} & {\small{}89.21} & \textbf{\small{}64.35} & {\small{}72.27} & {\small{}72.40} & \textbf{\small{}69.30} & \textbf{\small{}65.86} & {\small{}71.25} & \textbf{\small{}73.13} & \textbf{\small{}68.71} & \textbf{\small{}53.04}\tabularnewline
 & \emph{\small{}Last} & {\small{}94.08} & \textbf{\small{}95.50} & \textbf{\small{}92.98} & \textbf{\small{}92.84} & \textbf{\small{}94.00} & \textbf{\small{}92.27} & {\small{}87.23} & \textbf{\small{}61.07} & {\small{}71.84} & {\small{}72.03} & \textbf{\small{}69.30} & \textbf{\small{}65.69} & {\small{}71.16} & \textbf{\small{}72.37} & \textbf{\small{}68.42} & \textbf{\small{}52.95}\tabularnewline
\bottomrule 
\end{tabular}}
\end{table*}

\subsection{Extended results: WebVision}

Table \ref{tab:WebVision-1-1} extends the results obtained in
WebVision (first 50 classes) by including the performance in the last
training epoch when training with a ResNet-18 from scratch. Figure \ref{fig:WebVisionImages} 
shows examples of detected noisy images.
\begin{table*}[t]
\begin{centering}
\caption{\label{tab:WebVision-1-1}Top-1 accuracy in first 50 classes of WebVision.
We train all methods from scratch.}
\medskip{}
\par\end{centering}
\centering{}{\small{}}%
\begin{tabular}{ccccccccccc}
\toprule
 & {\small{}CE} & {\small{}FW \cite{2017_CVPR_ForwardLoss}} & {\small{}R \cite{2018_CVPR_JointOpt}} & {\small{}M \cite{2018_ICLR_mixup}} & {\small{}GCE \cite{2018_NeurIPS_GCE}} & {\small{}DB \cite{2019_ICML_BynamicBootstrapping}} & {\small{}DMI \cite{2019_NeurIPS_LDMI}} & {\small{}P\cite{2019_CVPR_JointOptimizImproved}} & {\small{}DM\cite{2020_ICLR_DivideMix}} & {\small{}DRPL (ours)}\tabularnewline
\hline 
\emph{\small{}Best} & {\small{}73.88} & {\small{}74.68} & {\small{}76.52} & {\small{}80.76} & {\small{}74.28} & {\small{}79.68} & {\small{}73.96} & {\small{}79.96} & {\small{}78.16} & \textbf{\small{}82.08}\tabularnewline
\emph{\small{}Last} & {\small{}73.76} & {\small{}74.32} & {\small{}76.24} & {\small{}79.96} & {\small{}74.08} & {\small{}79.56} & {\small{}73.88} & {\small{}79.44} & {\small{}76.84} & \textbf{\small{}82.00}\tabularnewline
\bottomrule 
\end{tabular}
\end{table*}
\begin{figure*}[t]
\centering{}\includegraphics[width=0.85\textwidth]{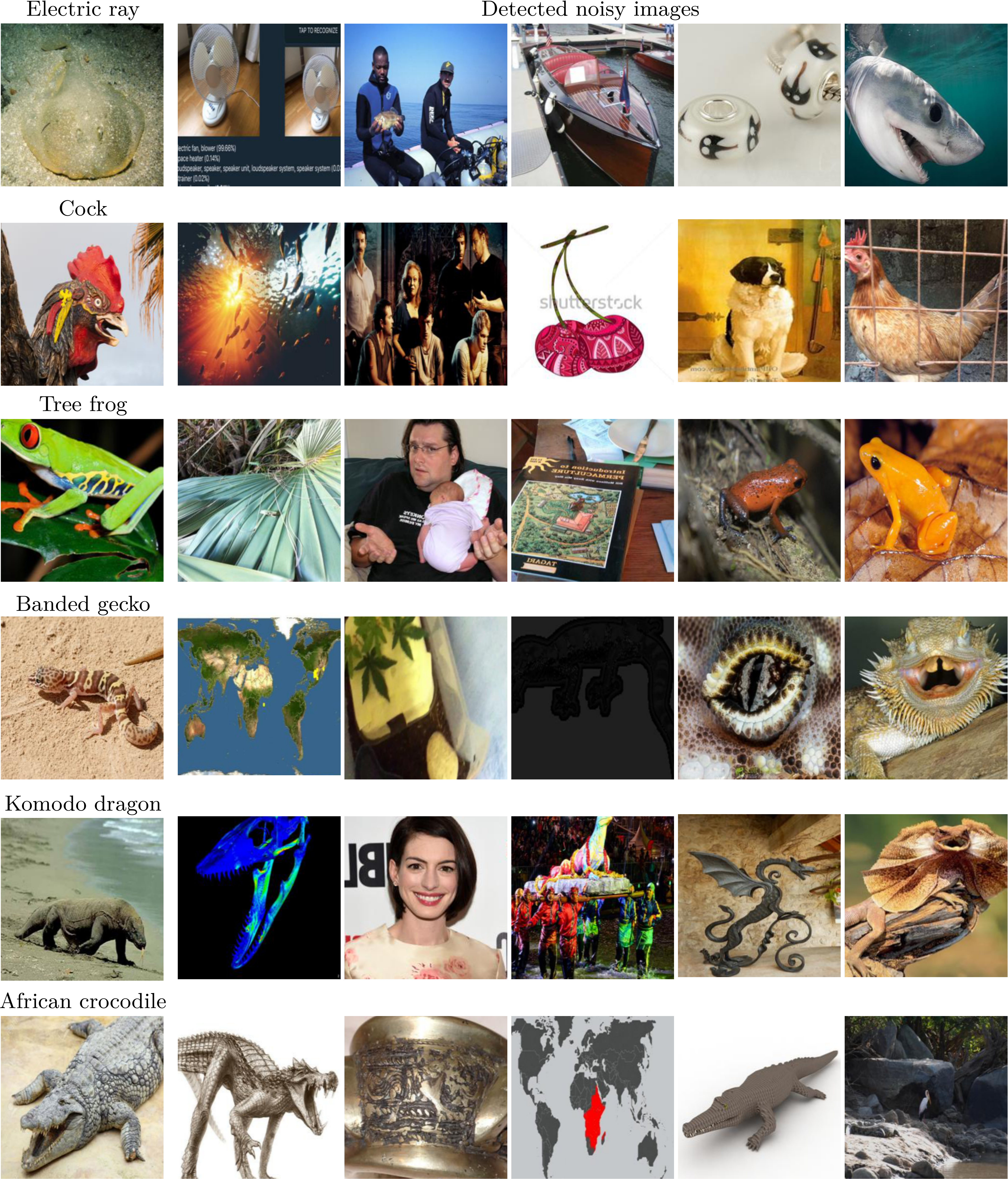}\caption{\label{fig:WebVisionImages} Detected noisy images in WebVision. First
column: example of clean image for the class. Second to sixth column:
detected noisy images.}
\end{figure*}

\end{document}